\documentclass[10pt,journal,compsoc,cspaper]{IEEEtran}
\usepackage{amsmath,amsfonts,amssymb}
\usepackage{algorithmic}
\usepackage{algorithm}
\usepackage{array}
\usepackage{booktabs}
\usepackage{multirow}
\usepackage{capt-of}
\usepackage{placeins}
\usepackage[caption=false,font=normalsize,labelfont=sf,textfont=sf]{subfig}
\usepackage{textcomp}
\usepackage{stfloats}
\usepackage{url}
\usepackage{graphicx}
\usepackage{cite}
\usepackage{xcolor}
\usepackage[hidelinks]{hyperref}
\begin{document}

\title{Smartphone Indoor Localization Enhanced by Hierarchical Multi-Sensor Context Recognition}

\author{Wenhui Feng\textsuperscript{1}, Muhan Weng\textsuperscript{1}, Jianyu Liu, Xiansheng Yang\textsuperscript{*}, Hao Zhang,
Liang Wang, and Yuan Zhuang,~\IEEEmembership{Senior Member,~IEEE}%
\thanks{Wenhui Feng, Muhan Weng, and Jianyu Liu are with the Electronic
Information School, Wuhan University, Wuhan, Hubei 430072, China
(e-mail: \{wenhuifeng, wengmuhan, jianyuliu\}@whu.edu.cn).}%
\thanks{Xiansheng Yang, Hao Zhang, and Yuan Zhuang are with the State Key
Laboratory of Information Engineering in Surveying, Mapping and Remote Sensing
(LIESMARS), Wuhan University, Wuhan, Hubei 430072, China
(e-mail: \{xiansheng.yang, 2021302191301, yuan.zhuang\}@whu.edu.cn).}%
\thanks{Liang Wang is with the Aerospace Information Research Institute,
Chinese Academy of Sciences, Beijing 100094, China
(e-mail: wangliang@aircas.ac.cn).}%
\thanks{\textsuperscript{*}Corresponding author: Xiansheng Yang.}%
\thanks{\textsuperscript{1}Wenhui Feng and Muhan Weng contributed equally to this work.}}

\markboth{Smartphone Indoor Localization Enhanced by Hierarchical Multi-Sensor Context Recognition}%
{Feng \MakeLowercase{\textit{et al.}}: Smartphone Indoor Localization}

\IEEEtitleabstractindextext{%
\begin{abstract}
Smartphone indoor localization requires low-cost, robust positioning across
scene transitions and phone-use variations. Pedestrian dead reckoning (PDR)
enables infrastructure-free deployment but accumulates errors from heading
drift, structural transitions, and carrying-mode changes. Existing methods often
treat contextual cues as a single event stream or local correction signal,
limiting their ability to address distinct localization errors. This paper
presents a framework enhanced by hierarchical multi-sensor context recognition.
A jointly data- and model-driven recognition model predicts environment, scene,
and activity probabilities, which are converted into scene-event and
action-context segments. Scene-event segments guide scene-constrained PDR
correction through corridor heading constraints, landmark attraction, fixed-lag
backtracking, and vertical-transition handling. Action-context segments drive
action-aware RoNIN-based position compensation for carrying-mode-dependent drift.
Experiments on cross-floor and multi-action trajectories against LiDAR-SLAM
references show that the proposed framework outperforms PDR and representative
learned inertial odometry baselines, demonstrating that hierarchical semantics
provide reliable, localization-specific correction cues.
\end{abstract}
}

\maketitle

\begin{figure*}[!t]
\centering
\includegraphics[width=\textwidth]{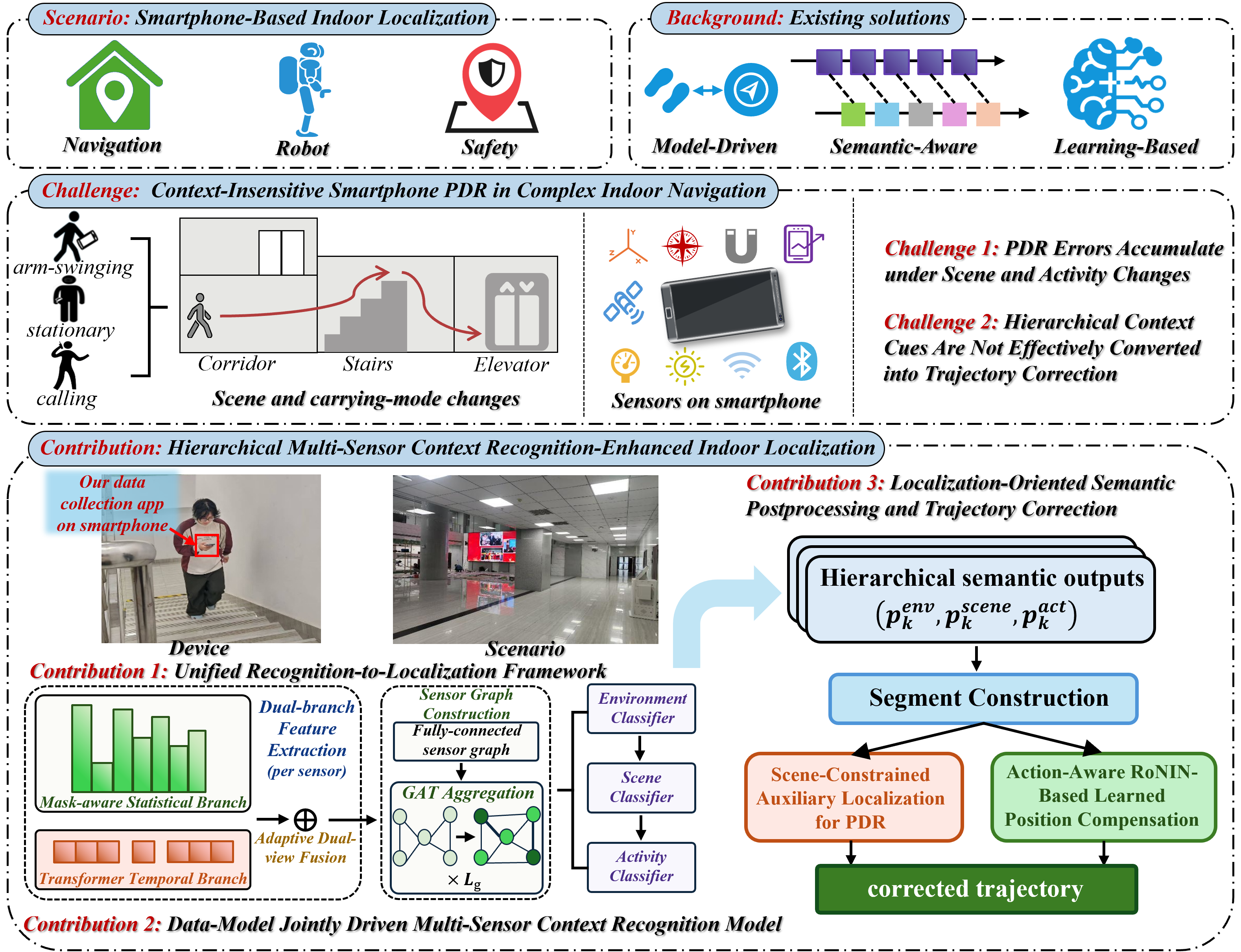}
\caption{Overview of the key challenges and the proposed smartphone localization
framework enhanced by hierarchical multi-sensor context recognition.}
\label{fig:overview}
\end{figure*}

\section{Introduction}
Pedestrian indoor localization and navigation are key enabling technologies for
smart buildings and smart communities, supporting location-aware services such as
indoor autonomous systems, personal guidance, emergency response, and facility
management. Outdoor positioning is widely supported by
global navigation satellite systems (GNSS), such as GPS and BeiDou. In indoor
and underground spaces, however, satellite signals are easily blocked by
physical obstructions, leading to severe attenuation. As a result,
high-precision GNSS services are difficult to extend to indoor and underground
scenes. Reliable, low-cost, and robust smartphone indoor localization remains an
important research problem \cite{reviewsmartphone,multisourcefusion}.

Various indoor localization techniques have been studied for this problem.
Wi-Fi and Bluetooth solutions are relatively easy to deploy. Their accuracy
and stability are affected by signal blockage, environmental change, and device
heterogeneity. Ultra-wideband, acoustic, and other infrastructure-assisted
methods can provide higher accuracy in controlled settings, but they introduce
deployment costs and maintenance burdens. Smartphone-based multimodal tracking
systems have fused radio, inertial, and magnetic-field observations
to improve indoor tracking robustness without dedicated user-side hardware
\cite{smartphone_multimodal_tracking,swiloc,smartphoneSLAM}. Inertial localization based on smartphone
sensors, especially pedestrian dead reckoning (PDR), is attractive because it is
self-contained, low-cost, and independent of external signal coverage. However,
PDR still suffers from step-length error, heading drift, and incorrect handling
of scene or floor transitions, causing the localization error to accumulate over
time \cite{smartpdr,reliablepdr,context3d,contextpersonal}.

Existing PDR and learned inertial localization methods usually rely on inertial
measurements, gait models, map constraints, structural priors, or data-driven
motion estimation \cite{activitysequence,semanticSLAM,activitysemantics,
activityloopclosure,ionet,ronin,tlio}. Although these designs address important
aspects of localization, they underuse hierarchical recognition of user context.
Indoor space is not homogeneous: corridors, stairs,
elevators, gates, and semi-open areas impose different trajectory constraints.
At the same time, phone-use modes such as holding, calling, arm-swinging, and
pocket placement change the observed inertial signals. Existing semantic-aided
methods often compress these cues into one event stream or use them only for
local correction, which limits robustness under complex scene transitions and
action changes.

The key issue is that different semantic layers can address different
localization error sources. Environment recognition can indicate indoor,
semi-open, and outdoor states for confidence filtering and transition handling.
Scene recognition can identify corridor, stair, elevator, gate, and related
structural events to trigger heading constraints, landmark attraction,
vertical-transition handling, and backtracking smoothing. Activity recognition
can identify phone
holding, calling, arm-swinging, and pocket modes to compensate
carrying-mode-dependent inertial drift. In this way, context recognition can
convert classification outputs into correction cues that directly enhance
trajectory estimation.

This paper presents a smartphone indoor localization framework enhanced by
hierarchical multi-sensor context recognition. Multi-sensor temporal windows
are first processed by a data-model jointly driven context recognition model to
obtain environment-, scene-, and activity-level probabilities. The predicted
semantic outputs are then converted
into scene-event and action-context segments through a localization-oriented
postprocessing interface. Scene events guide a scene-constrained PDR branch,
while action context guides an action-aware RoNIN compensation branch.

The main contributions of this paper are summarized as follows:

\noindent\hangindent=1.5em\hangafter=1
1) A smartphone localization framework enhanced by hierarchical multi-sensor
context recognition that maps environment, scene, and activity recognition
outputs to localization enhancement modules, enabling unified modeling from
recognition results to trajectory correction.

\noindent\hangindent=1.5em\hangafter=1
2) A data-model jointly driven multi-sensor context recognition
model that fuses temporal, statistical, and cross-sensor representations to
predict hierarchical semantic probabilities for localization enhancement.

\noindent\hangindent=1.5em\hangafter=1
3) A localization-oriented semantic postprocessing and trajectory correction
mechanism, including scene-event extraction, confidence filtering, structural
constraint triggering, vertical-transition handling, action-context compensation,
and backtracking smoothing, with its effectiveness verified through multi-route
experiments.

\begin{table*}[!t]
\centering
\footnotesize
\caption{Context Cues in Related Localization Methods}
\label{tab:related_work_comparison}
\setlength{\tabcolsep}{4pt}
\renewcommand{\arraystretch}{1.14}
\begin{tabular}{@{}>{\raggedright\arraybackslash}p{0.18\textwidth}
>{\raggedright\arraybackslash}p{0.45\textwidth}
>{\raggedright\arraybackslash}p{0.31\textwidth}@{}}
\toprule
Method family & Context cue and use & Our Work \\
\midrule
Context-aware PDR & Motion state, carrying mode, or map structure adjusts steps, headings, and floor transitions. & Scene events drive map-guided PDR correction. \\
Semantic-aided localization & Scene, activity, or landmark events constrain routes, topology, or graph factors. & Environment, scene, and activity cues are separated before localization. \\
Learned inertial odometry & IMU windows predict velocity or displacement in a learned trajectory stream. & Action context compensates RoNIN trajectory drift. \\
\bottomrule
\end{tabular}
\end{table*}

\section{Related Work}
Smartphone indoor localization has evolved from standalone PDR to hybrid
systems that combine inertial sensing, structural context, wireless cues, and
learned motion estimation \cite{reviewsmartphone,multisourcefusion}. Classical
PDR estimates pedestrian displacement from step detection, step length, and
heading updates, and remains attractive because of its low cost and
interpretability \cite{smartpdr}. Later systems improve robustness by adding
motion-state adaptation, personalized parameters, floor-transition handling, or
map structure \cite{reliablepdr,context3d,contextpersonal,multifloor,
realtime3d,pylon}. These context-aware PDR methods can reduce drift when the motion
model and structural prior are reliable. However, their correction rules are
usually designed for a single dead-reckoning chain, making such contextual cues
difficult to reuse in learned inertial odometry or other localization schemes.
Beyond smartphone PDR, recent indoor positioning studies have explored
RIS-assisted radio environments, learning-based multistatic UWB radar using
low-cost devices, and site-survey-free positioning based on RSSI diversity
\cite{rispositioning,uwbradar,optimaps}. These studies broaden the available
measurement and infrastructure choices for robust indoor positioning.
Construction-oriented studies further emphasize deployment considerations by
comparing on-site indoor positioning technologies and applying BLE-based RTLS
to worker proximity alerts \cite{li2020top10,huang2021proximity}.

Semantic-aided localization introduces higher-level information into the
positioning process. Scene and activity recognition methods infer walking,
stairs, elevators, and phone-use modes from smartphone sensors
\cite{scenerec,activitycnn,positionawarehar}. Recent sensor-fusion activity
recognition studies also show that heterogeneous time-series representations
and self-supervised multimodal representations can improve wearable and mobile
recognition robustness \cite{qin2020har_fusion,refuseact}.
Construction monitoring research has also reviewed sensor-based mapping, scene
understanding, positioning, and tracking, while a recent indoor worker-monitoring
system fused BLE location, body orientation, and accelerometer-derived
productivity states \cite{rao2022monitoring,khazen2024workers}.
Beyond inertial sensing, a quality-driven geomagnetic framework quantifies
fingerprint distinguishability and combines constrained trajectory matching
with learned feature comparison to improve positioning robustness
\cite{geomagneticquality}. FINLOS-TriSeg integrates fisheye semantic masks with
satellite geometry for GNSS NLOS identification, illustrating how complementary
semantic and geometric cues can improve environmental perception reliability
\cite{finlostriseg}. These findings suggest that incorporating additional
modalities and explicit measurement-quality cues into hierarchical context
recognition could further improve recognition accuracy and localization
robustness. The recognized labels
can then be used as landmarks, sequence constraints, topology cues, or graph
factors to support position inference and drift correction
\cite{activitysemanticdesc,activitysequence,semanticSLAM,activitysemantics,
activityloopclosure}. These studies show that semantic events are useful for
localization, especially when geometric measurements alone are weak. Still, many
methods merge different semantic labels into a single time-ordered event
sequence. This makes it difficult to distinguish structural scene cues, such as
corridors or stairs, from action and carrying-mode cues, which mainly affect
inertial motion estimation.

Learning-based inertial odometry estimates motion directly from IMU windows and
reduces the dependence on hand-crafted gait assumptions. RIDI and IONet first
showed that data-driven models can suppress double-integration drift
\cite{ridi,ionet}. RoNIN improved robustness under unconstrained daily motion,
while IDOL, TLIO, LLIO, and MEIO further considered orientation estimation,
tight filtering, lightweight deployment, and adaptive computation
\cite{ronin,idol,tlio,llio,meionet,watchposition}. These methods provide competitive trajectory
baselines based primarily on inertial sequences and cover complementary design
families, including double-integration learning, robust neural inertial
navigation, filter-coupled learned odometry, and efficient motion-aware inertial
odometry. They usually do not expose an explicit
interface through which scene structure or phone-use semantics can be used as
external correction cues.

Our work uses a hierarchical semantic interface to provide semantic information
for different localization backends. Environment and scene outputs are converted
into scene-event segments for scene-constrained PDR correction, while activity
outputs are converted into action-context segments for action-aware RoNIN
compensation.
Table~\ref{tab:related_work_comparison} summarizes this distinction.

\begin{figure*}[!t]
\centering
\includegraphics[width=\textwidth]{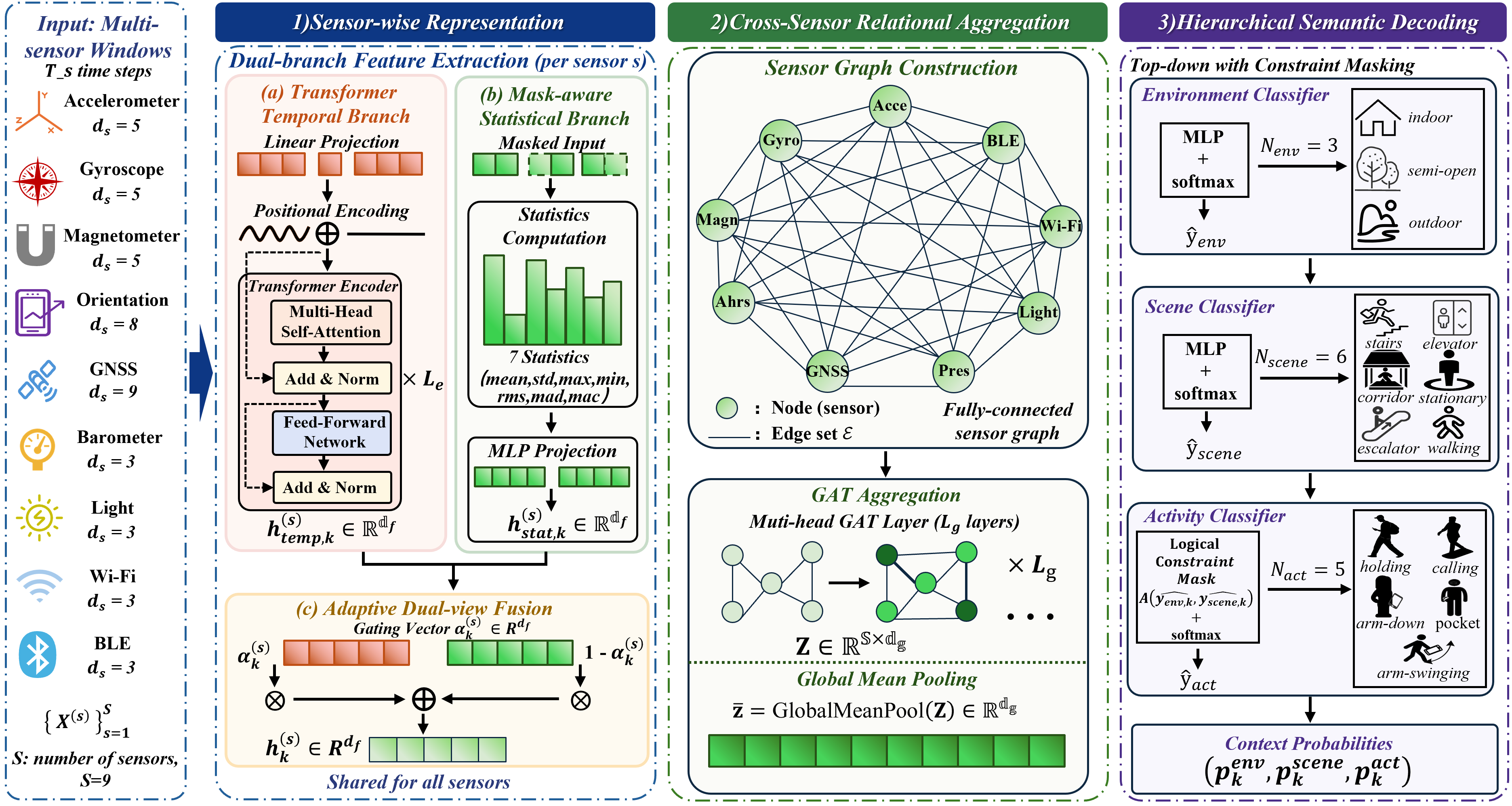}
\caption{Hierarchical multi-sensor context recognition model. The model
extracts temporal and mask-aware statistical features for each sensor, aggregates
cross-sensor relations with a fully connected graph-attention module, and
decodes environment, scene, and activity probabilities with top-down semantic
constraints.}
\label{fig:recognition_unified_framework}
\end{figure*}
\section[Hierarchical Multi-Sensor Context Recognition Model]{Hierarchical Multi-Sensor Context\\Recognition Model}

\subsection{Hierarchical Context Semantics}

The recognition model converts heterogeneous smartphone observations into
three hierarchical semantic layers. This separation preserves the context cues
needed by downstream localization correction at the environment, scene, and
activity levels.

For the $k$-th temporal window, the multi-sensor input is denoted by

\begin{equation}
\mathcal{X}_k=\left\{\mathbf{X}^{(s)}_k,\mathbf{M}^{(s)}_k\right\}_{s=1}^{S},
\label{eq:input_window_set}
\end{equation}
where $S$ is the number of sensor modalities, $\mathbf{X}^{(s)}_k\in
\mathbb{R}^{T_s\times d_s}$ is the feature matrix of sensor $s$ in that
window, and $\mathbf{M}^{(s)}_k\in\{0,1\}^{T_s}$ is the validity mask used to
identify available samples under asynchronous sensing or padding. The model
outputs three semantic layers:
\begin{equation}
\mathbf{p}^{\mathrm{env}}_k,\quad
\mathbf{p}^{\mathrm{scene}}_k,\quad
\mathbf{p}^{\mathrm{act}}_k,
\label{eq:three_level_outputs}
\end{equation}
which represent environment, scene, and activity probabilities. The environment
layer distinguishes indoor, semi-open, and outdoor exposure states. The scene
layer describes navigation states such as corridor, stairs, elevator, escalator,
walking, and stationary. The activity layer describes phone-use and carrying
modes, including holding, calling, arm-down, arm-swinging, and pocket.
Accordingly, the output dimensions are $N_{\mathrm{env}}=3$,
$N_{\mathrm{scene}}=6$, and $N_{\mathrm{act}}=5$. The localization modules use
these layers differently: scene-related predictions trigger structural PDR
correction, while activity predictions condition the learned RoNIN
compensation branch.

\subsection{Data-Model Jointly Driven Multi-Sensor Context Recognition Model}

This semantic front end connects heterogeneous smartphone sensing with
localization correction. Scene structure and phone-use mode affect different
localization error sources: structural events can activate map-aware PDR
constraints, while carrying-mode changes can condition RoNIN compensation. The
recognition model turns asynchronous multi-sensor windows into reliable context
probabilities, labels, and confidence scores for the downstream localization
modules.
Fig.~\ref{fig:recognition_unified_framework} summarizes this model. It follows
three stages: sensor-wise representation learning, cross-sensor relational
aggregation, and hierarchical semantic decoding.

\subsubsection{Sensor-Wise Temporal Encoding}

Each sensor stream is first processed by a temporal branch to capture motion
patterns within the current window. For sensor $s$, the raw feature sequence is
projected into a shared latent space and combined with positional encoding:
\begin{equation}
\mathbf{U}^{(s)}_k=
\mathrm{PE}
\!\left(
\mathbf{X}^{(s)}_k\mathbf{W}^{(s)}_{\mathrm{in}}
\right),
\label{eq:temporal_input}
\end{equation}
where $\mathbf{W}^{(s)}_{\mathrm{in}}$ is the input projection matrix and
$\mathrm{PE}(\cdot)$ denotes positional encoding. The Transformer encoder then
uses the validity mask as the padding mask, with $L_e$ denoting the number of
encoder layers:
\begin{equation}
\mathbf{H}^{(s)}_k=
\mathrm{Enc}^{(s)}
\!\left(
\mathbf{U}^{(s)}_k;\mathbf{M}^{(s)}_k
\right),
\label{eq:temporal_encoder}
\end{equation}
where invalid samples neither attend to nor are attended by valid samples. A
compact temporal embedding is obtained by masked average pooling:
\begin{equation}
\mathbf{h}^{(s)}_{\mathrm{temp},k}
=
\frac{
\sum_{t=1}^{T_s}M^{(s)}_{k,t}\mathbf{H}^{(s)}_{k,t}
}{
\max\!\left(\sum_{t=1}^{T_s}M^{(s)}_{k,t},1\right)
}.
\label{eq:temporal_pooling}
\end{equation}
The resulting embedding satisfies
$\mathbf{h}^{(s)}_{\mathrm{temp},k}\in\mathbb{R}^{d_f}$. If a
sensor-window contains no valid sample, the temporal embedding is set to zero
and the fusion gate relies on the available modalities. This branch mainly
describes step rhythm, local transitions, and longer-range dependencies
embedded in inertial and environmental signals.

\subsubsection{Mask-Aware Statistical Encoding}

Heterogeneous smartphone sensing also requires stable window-level statistics,
especially for sparse or intermittently available modalities. The valid index
set of sensor $s$ in the $k$-th window is
\begin{equation}
\mathcal{T}^{(s)}_k=\{t\mid M^{(s)}_{k,t}=1\}.
\label{eq:valid_index_set}
\end{equation}
For each feature channel $f$, the statistical branch computes
\begin{equation}
\boldsymbol{\phi}^{(s)}_{k,f}
=
\left[
\mu_f,\sigma_f,x_f^{\max},x_f^{\min},
\mathrm{rms}_f,\mathrm{mad}_f,\mathrm{mac}_f
\right],
\label{eq:stat_channel_descriptor}
\end{equation}
where the statistics are evaluated only on $\mathcal{T}^{(s)}_k$, and
$\mathrm{rms}_f$, $\mathrm{mad}_f$, and $\mathrm{mac}_f$ denote root mean
square, mean absolute deviation, and mean absolute change over consecutive
valid pairs, respectively. The sensor-level descriptor stacks all
channel-wise descriptors:
\begin{equation}
\boldsymbol{\phi}^{(s)}_k=
\mathrm{Concat}
\!\left(
\left\{
\boldsymbol{\phi}^{(s)}_{k,f}
\right\}_{f=1}^{d_s}
\right).
\label{eq:stat_descriptor}
\end{equation}
If a valid set is empty, the corresponding statistic is set to zero before
projection. The descriptor is mapped to the shared feature space through an MLP:
\begin{equation}
\mathbf{h}^{(s)}_{\mathrm{stat},k}=
\mathrm{MLP}^{(s)}_{\mathrm{stat}}
\!\left(
\boldsymbol{\phi}^{(s)}_k
\right).
\label{eq:stat_project}
\end{equation}
The resulting statistical embedding satisfies
$\mathbf{h}^{(s)}_{\mathrm{stat},k}\in\mathbb{R}^{d_f}$.
This branch supplies robust distributional evidence and reduces the influence
of padding, missing samples, and short-term fluctuations.

\subsubsection{Adaptive Dual-View Fusion}

The temporal and statistical views contribute differently under different
contexts. A learnable gating mechanism balances their
contributions:
\begin{equation}
\boldsymbol{\alpha}^{(s)}_k=
\sigma
\!\left(
\mathbf{W}_{\alpha}
\left[
\mathbf{h}^{(s)}_{\mathrm{temp},k}\,\|\,\mathbf{h}^{(s)}_{\mathrm{stat},k}
\right]
\!+\mathbf{b}_{\alpha}
\right),
\label{eq:fusion_gate}
\end{equation}
\begin{equation}
\mathbf{h}^{(s)}_k=
\boldsymbol{\alpha}^{(s)}_k\odot\mathbf{h}^{(s)}_{\mathrm{temp},k}
+
\left(
1-\boldsymbol{\alpha}^{(s)}_k
\right)\odot\mathbf{h}^{(s)}_{\mathrm{stat},k},
\label{eq:fusion_result}
\end{equation}
where $\boldsymbol{\alpha}^{(s)}_k\in(0,1)^{d_f}$ is the fusion gate,
$\mathbf{h}^{(s)}_k\in\mathbb{R}^{d_f}$ is the fused sensor-wise embedding,
$\odot$ denotes element-wise multiplication, and $\|\,$ denotes vector
concatenation. This design allows the module to favor temporal evidence when
the window contains distinctive motion evolution, while relying more heavily on
stable statistics when the signal is noisy or weakly structured.

\subsubsection{Cross-Sensor Relational Aggregation}

After dual-view fusion, one node embedding is obtained for each sensor. Because
scene and activity cues may be distributed across complementary sensor responses,
a fully connected sensor graph models inter-sensor relations through stacked
graph-attention layers. With
$\mathbf{z}^{(0)}_{s,k}=\mathbf{h}^{(s)}_k$, the final node matrix is
$\mathbf{Z}_k=[\mathbf{z}^{(L_g)}_{1,k},\ldots,
\mathbf{z}^{(L_g)}_{S,k}]^\top\in\mathbb{R}^{S\times d_g}$. The window-level
representation is obtained by global mean pooling over the final sensor nodes:
\begin{equation}
\bar{\mathbf{z}}_k=
\mathrm{GlobalMeanPool}
\!\left(
\mathbf{Z}_k
\right),
\label{eq:graph_pooling}
\end{equation}
with $\bar{\mathbf{z}}_k\in\mathbb{R}^{d_g}$ and $L_g$ denoting the number of
graph-attention layers. This stage captures cross-modal relations among
inertial, orientation, pressure, light, and wireless cues.

\subsubsection{Hierarchical Semantic Decoding}

The final stage uses hierarchical semantic constraints to decode the aggregated
graph representation in a top-down order across environment, scene, and
activity levels. The environment prediction is first obtained from the global
feature:
\begin{equation}
\mathbf{p}^{\mathrm{env}}_k=
\mathrm{softmax}
\!\left(
f_{\mathrm{env}}(\bar{\mathbf{z}}_k)
\right).
\label{eq:env_head}
\end{equation}
The scene prediction is then conditioned on the same global feature together
with the environment output:
\begin{equation}
\mathbf{p}^{\mathrm{scene}}_k=
\mathrm{softmax}
\!\left(
f_{\mathrm{scene}}
\!\left(
\left[
\bar{\mathbf{z}}_k\,\|\,\mathbf{p}^{\mathrm{env}}_k
\right]
\right)
\right).
\label{eq:scene_head}
\end{equation}
For windows assigned to an outdoor environment in the training data, the scene
target is marked as unavailable. At inference, an N/A scene state is generated
only in postprocessing when the environment prediction indicates outdoor motion.
Finally, the activity decoder receives the global feature and both upper-level
semantic outputs:
\begin{equation}
\boldsymbol{\ell}^{\mathrm{act}}_k=
f_{\mathrm{act}}
\!\left(
\left[
\bar{\mathbf{z}}_k\,\|\,\mathbf{p}^{\mathrm{env}}_k\,\|\,\mathbf{p}^{\mathrm{scene}}_k
\right]
\right).
\label{eq:act_logit}
\end{equation}
During both training and normal inference, the activity decoder is conditioned
on the soft upper-level probabilities. A hard semantic filter is used only as a
low-confidence consistency postprocessing step. When this filter is enabled,
inadmissible activity classes are suppressed according to the predicted
environment-scene pair:
\begin{equation}
\tilde{\ell}^{\mathrm{act}}_{k,c}=
\begin{cases}
\ell^{\mathrm{act}}_{k,c}, & c\in A(\hat{y}^{\mathrm{env}}_k,\hat{y}^{\mathrm{scene}}_k),\\
-\gamma, & \text{otherwise},
\end{cases}
\label{eq:act_mask}
\end{equation}
where $\hat{y}^{\mathrm{env}}_k=\arg\max_c \mathbf{p}^{\mathrm{env}}_{k,c}$ and
$\hat{y}^{\mathrm{scene}}_k=\arg\max_c \mathbf{p}^{\mathrm{scene}}_{k,c}$ are
the hard environment and scene predictions, $A(\cdot,\cdot)$ is the
admissible activity set, and $\gamma$ is a large positive constant. If the
filter is not activated, $\tilde{\boldsymbol{\ell}}^{\mathrm{act}}_k=
\boldsymbol{\ell}^{\mathrm{act}}_k$. The activity probability is then produced
by
\begin{equation}
\mathbf{p}^{\mathrm{act}}_k=
\mathrm{softmax}
\!\left(
\tilde{\boldsymbol{\ell}}^{\mathrm{act}}_k
\right).
\label{eq:act_head}
\end{equation}
This top-down decoder preserves coarse-to-fine semantics and avoids
physically inconsistent context combinations.

The recognition front-end is trained with a compact multi-task objective. Let
$y_k^{\mathrm{env}}$, $y_k^{\mathrm{scene}}$, and $y_k^{\mathrm{act}}$ denote
the three labels, and let $m_k^{\mathrm{scene}}\in\{0,1\}$ indicate whether the
scene label is available. The total loss is
\begin{equation}
\mathcal{L}_{\mathrm{rec}}
=
\mathcal{L}_{\mathrm{env}}
+0.8\,\mathcal{L}_{\mathrm{scene}}
+0.8\,\mathcal{L}_{\mathrm{act}},
\label{eq:rec_loss}
\end{equation}
where $\mathcal{L}_{\mathrm{env}}$ and $\mathcal{L}_{\mathrm{act}}$ are
class-weighted cross-entropy losses, and
\begin{equation}
\mathcal{L}_{\mathrm{scene}}
=
\frac{
\sum_k m_k^{\mathrm{scene}}
\mathrm{CE}\!\left(
\mathbf{p}^{\mathrm{scene}}_k,
y_k^{\mathrm{scene}}
\right)
}{
\max\!\left(\sum_k m_k^{\mathrm{scene}},1\right)
}.
\label{eq:scene_masked_loss}
\end{equation}
With this masking strategy, outdoor or otherwise unavailable scene labels do not contribute to the
scene loss, while the environment and activity branches remain supervised.

\subsection{Recognition-to-Localization Interface}

The semantic outputs of the hierarchical context recognition model are organized
into continuous, temporally aligned cues for downstream localization refinement.

A window-level prediction record is defined as
\begin{equation}
o_k=
\left\{
t_k^{\mathrm{st}},t_k^{\mathrm{ed}},
\mathbf{p}^{\mathrm{env}}_k,
\mathbf{p}^{\mathrm{scene}}_k,
\mathbf{p}^{\mathrm{act}}_k
\right\},
\label{eq:window_record}
\end{equation}
where $t_k^{\mathrm{st}}$ and $t_k^{\mathrm{ed}}$ are the starting and ending
timestamps of the $k$-th window. For each semantic level
$r\in\{\mathrm{env},\mathrm{scene},\mathrm{act}\}$, the window-level dominant
label and confidence are
\begin{equation}
\hat{c}^{r}_k=\arg\max_c p^{r}_{k,c},\quad
\rho^{r}_k=\max_c p^{r}_{k,c}.
\label{eq:window_label_conf}
\end{equation}
Window records are sorted by timestamp. If two records overlap, the overlapped
interval is assigned to the record with the larger confidence at the semantic
level being processed. Adjacent records are merged when they have the same
dominant label and their temporal gap is no larger than $\tau_g$.

\begin{figure*}[!t]
\centering
\includegraphics[width=\textwidth]{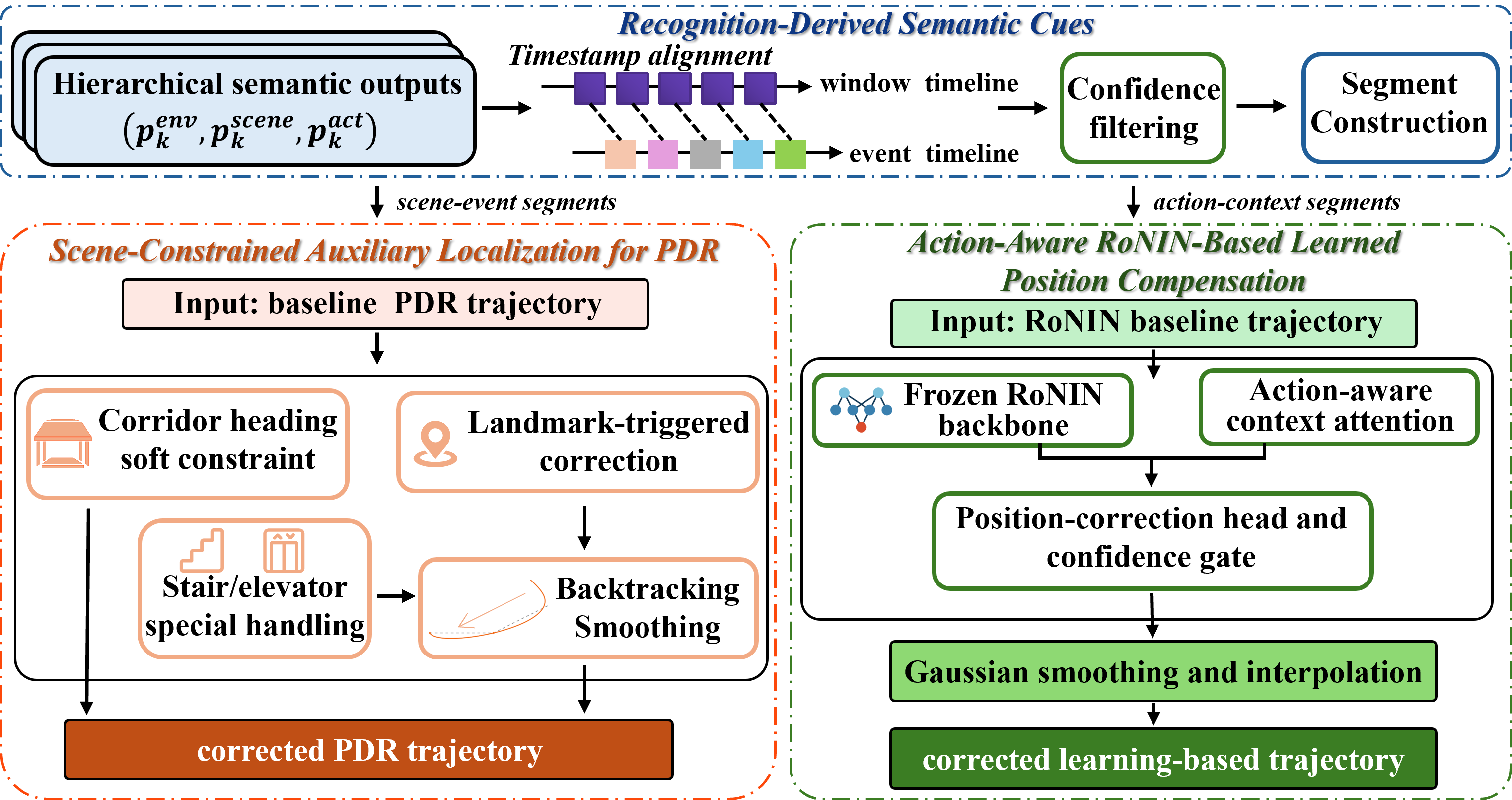}
\caption{Auxiliary localization framework driven by recognition-derived semantic
cues. Scene-event segments activate scene-constrained auxiliary localization for
PDR, while action-context segments guide action-aware RoNIN-based learned
position compensation.}
\label{fig:auxiliary_localization_framework}
\end{figure*}

The first interface is a set of scene-event segments,
\begin{equation}
e_m=
\left\{
t_m^{\mathrm{st}},t_m^{\mathrm{ed}},
\hat{c}^{\mathrm{env}}_m,
\hat{c}^{\mathrm{scene}}_m,
q_m
\right\},
\label{eq:scene_segment}
\end{equation}
where $\hat{c}^{\mathrm{env}}_m$ and $\hat{c}^{\mathrm{scene}}_m$ are the
dominant environment and scene labels in the segment and $q_m$ denotes its
segment-level confidence. Let $\mathcal{I}_m$ denote the set of windows merged
into $e_m$. For indoor scene events, the segment confidence is computed as
\begin{equation}
q_m=
\frac{1}{|\mathcal{I}_m|}
\sum_{k\in\mathcal{I}_m}
\min\!\left(\rho^{\mathrm{env}}_k,\rho^{\mathrm{scene}}_k\right),
\label{eq:scene_segment_conf}
\end{equation}
Outdoor-transition segments use the average environment confidence only, as the
scene label is unavailable. The environment label is mainly used to
identify indoor-outdoor transitions, while the scene label supports corridor
constraints and vertical-transition handling. These segments are supplied to the
scene-constrained localization path to activate the corridor heading soft
constraint, entrance constraints, and vertical-transition handling.

The second interface is a set of action-context segments,
\begin{equation}
a_n=
\left\{
t_n^{\mathrm{st}},t_n^{\mathrm{ed}},
\hat{c}^{\mathrm{act}}_n,
\bar{\mathbf{p}}^{\mathrm{act}}_n
\right\},
\label{eq:action_segment}
\end{equation}
where $\hat{c}^{\mathrm{act}}_n$ is the dominant activity class and
$\bar{\mathbf{p}}^{\mathrm{act}}_n$ is the segment-level action-probability
vector. If $\mathcal{J}_n$ is the set of windows in $a_n$, then
\begin{equation}
\bar{\mathbf{p}}^{\mathrm{act}}_n
=
\frac{1}{|\mathcal{J}_n|}
\sum_{k\in\mathcal{J}_n}
\mathbf{p}^{\mathrm{act}}_k,\quad
\hat{c}^{\mathrm{act}}_n=\arg\max_c \bar{p}^{\mathrm{act}}_{n,c}.
\label{eq:action_segment_avg}
\end{equation}
These action segments are used by the action-aware compensation path to
modulate correction strength according to the carrying mode and motion pattern.
The two segment types form the inputs to the localization branches.

\section{Auxiliary Localization Methods}

The proposed semantic interface can support localization backends with
different error mechanisms. In conventional PDR, structural scene cues can
constrain accumulated heading and transition errors through map-aware
correction. In learned inertial odometry, phone-use and carrying-mode cues can
refine motion estimation when the inertial feature distribution changes. The
auxiliary localization framework contains two branches: a scene-constrained
auxiliary localization branch for PDR and an action-aware RoNIN-based learned
position compensation branch, as summarized in
Fig.~\ref{fig:auxiliary_localization_framework}.

\subsection{Scene-Constrained Auxiliary Localization for PDR}

The first path starts from a baseline step-wise PDR trajectory and uses
scene-event segments to activate corridor, landmark, stair, and elevator
corrections. Only reliable segments are retained:
\begin{equation}
\mathcal{E}_{\mathrm{valid}}
=
\left\{
e_m
\;\middle|\;
q_m\ge\tau_q,\;
N_m\ge N_{\min}(\hat{c}^{\mathrm{scene}}_m)
\right\},
\label{eq:iv_valid_events}
\end{equation}
where $\tau_q$ is the confidence threshold, $N_m$ is the number of aligned
windows contained in segment $e_m$, and $N_{\min}(\cdot)$ is the minimum
window-count requirement. Stair, elevator, and outdoor-transition events are
required to persist over at least two consecutive windows to suppress isolated
false alarms.

The baseline PDR first detects steps from acceleration-magnitude peaks with a
minimum step-interval constraint. Let $t_i$ denote the timestamp of the $i$-th
detected step, and let $A_i^{\max}$ and $A_i^{\min}$ be the local
acceleration-magnitude extrema around the detected peak. The raw step length
follows the Weinberg model:
\begin{equation}
s_i^0
=
K_w
\left(
A_i^{\max}-A_i^{\min}
\right)^{1/4},
\label{eq:iv_weinberg}
\end{equation}
where $K_w$ is the calibrated step-length coefficient. For flat-ground walking,
$s_i=s_i^0$; for validated stair segments, the stride is shortened by a stair
factor $\eta_{\mathrm{stair}}$.

The raw step heading $\psi_i^0$ is obtained from the AHRS yaw after initial
alignment with the floor-plan coordinate system. The starting point, yaw offset,
principal corridor directions, and landmark positions are defined in the same
map frame; accordingly, this branch is map-guided.
Let $\mathbf{p}_i=[x_i,y_i]^\top$ denote the position after the step, $s_i$ the
scene-adjusted step length, and $\bar{\psi}_i$ the corrected heading used for
integration. The nominal update is
\begin{equation}
\mathbf{p}_i
=
\mathbf{p}_{i-1}
+
s_i
\begin{bmatrix}
\cos\bar{\psi}_i\\
\sin\bar{\psi}_i
\end{bmatrix}.
\label{eq:iv_pdr_update}
\end{equation}
Before any semantic correction, $\bar{\psi}_i=\psi_i^0$.

The main correction mechanism in ordinary indoor walking is the corridor
heading soft constraint. The method uses a set of principal corridor
directions defined in the map coordinate system. For a step detected inside a
reliable indoor segment, the nearest principal direction is selected by
\begin{equation}
d_i^\star
=
\arg\min_{d\in\mathcal{D}_{\mathrm{corr}}}
\left|
\Delta(\psi_i^0,d)
\right|,
\label{eq:iv_corridor_target}
\end{equation}
where $\mathcal{D}_{\mathrm{corr}}$ is the map-defined corridor direction set
and $\Delta(\cdot,\cdot)$ denotes the wrapped angular difference. If the
deviation remains within a tolerance band, the heading is softly pulled toward
the nearest principal direction:
\begin{equation}
\bar{\psi}_i
=
\psi_i^0-\lambda\,\Delta(\psi_i^0,d_i^\star),
\quad
\left|
\Delta(\psi_i^0,d_i^\star)
\right|<\varepsilon,
\label{eq:iv_corridor_soft}
\end{equation}
where $\lambda\in(0,1)$ is the pull strength and $\varepsilon$ is the corridor
tolerance. This design reduces heading drift while preserving local trajectory
flexibility. The corridor direction set $\mathcal{D}_{\mathrm{corr}}$ and the
two correction parameters are selected based on the floor-plan geometry and the
target deployment environment.

Beyond corridor regularization, scene-event segments also trigger
landmark-related corrections. Outdoor gate transitions, stair entrances or
exits, and elevator entrances are each associated with a predefined landmark
position in the map coordinate system. A landmark correction is activated only
when both temporal and spatial gating conditions are satisfied:
\begin{equation}
t_i\in
\left[
t_m^{\mathrm{st}}-T_m^{\mathrm{pre}},
t_m^{\mathrm{ed}}+T_m^{\mathrm{post}}
\right],
\quad
\left\|
\mathbf{p}_i-\mathbf{c}_m
\right\|_2<\tau_m,
\label{eq:iv_landmark_gate}
\end{equation}
where $\mathbf{c}_m$ is the landmark associated with scene event $e_m$,
$T_m^{\mathrm{pre}}$ and $T_m^{\mathrm{post}}$ are event-dependent temporal
margins, and $\tau_m$ is the corresponding spatial activation radius. This
design allows the system to anticipate a gate or stair entrance slightly
before the exact semantic boundary and also prevents distant false triggers
from abruptly dragging the track toward an unrelated landmark.

Once a landmark correction is triggered, a backtracking smoothing mechanism
pulls the current buffered segment toward the mapped target and suppresses
abrupt breaks in the polyline trajectory. Suppose that the current buffered
segment spans from time
$t_s$ to $t_i$, and that the current endpoint $\mathbf{p}_i$ is to be corrected
to a target landmark $\mathbf{c}$. The endpoint drift is written as
\begin{equation}
\boldsymbol{\delta}_i=\mathbf{c}-\mathbf{p}_i.
\label{eq:iv_drift}
\end{equation}
For each buffered point $\mathbf{p}_j$ with timestamp $t_j\in[t_s,t_i]$, the
smoothed backtracking update is
\begin{equation}
\mathbf{p}_j'
=
\mathbf{p}_j
+
\omega_j\boldsymbol{\delta}_i,
\quad
\omega_j=
\frac{t_j-t_s}{t_i-t_s}.
\label{eq:iv_backtracking}
\end{equation}
Earlier points receive smaller corrections, while points closer to the detected
event absorb a larger share of the drift. This gradual update
preserves trajectory continuity and better matches the physical behavior of a
user approaching a doorway, stair entrance, or elevator landing. The correction
is implemented with a fixed-lag buffer: only the current uncommitted segment is
modified before it is appended to the final trajectory. In online use this
introduces a bounded output delay, while in offline evaluation the same buffered
update is applied as postprocessing to the corresponding segment.

Elevator events require additional handling because they involve waiting,
entering, and short stationary periods that are poorly described by ordinary
walking PDR. The method monitors short consecutive high-confidence elevator
detections before switching to elevator mode. Once activated, the
recent trajectory is retrospectively corrected to a waiting position
$\mathbf{c}_{\mathrm{wait}}$ and then to an inside-elevator position
$\mathbf{c}_{\mathrm{in}}$. For a buffered point with timestamp $t_j$, the
retrospective update can be summarized as
\begin{equation}
\mathbf{p}_j'=
\begin{cases}
\mathbf{c}_{\mathrm{wait}}, & t_i-t_j>T_{\mathrm{enter}},\\
\left(1-\xi_j\right)\mathbf{c}_{\mathrm{wait}}
+
\xi_j\mathbf{c}_{\mathrm{in}}, & 0\le t_i-t_j\le T_{\mathrm{enter}},
\end{cases}
\label{eq:iv_elevator_backtracking}
\end{equation}
where
\begin{equation}
\xi_j=
1-\frac{t_i-t_j}{T_{\mathrm{enter}}}
\label{eq:iv_elevator_ratio}
\end{equation}
is the normalized entering progress and $T_{\mathrm{enter}}$ is the entering
duration. After the backtracking update, the trajectory is temporarily locked
to the elevator state until the validated elevator interval ends. Together with
stair-specific stride scaling, gate or stair entrance attraction, and the
corridor heading soft constraint, this path yields a corrected PDR trajectory
that remains compatible with both the floor-plan structure and the recognized
scene transitions.

\subsection{Action-Aware RoNIN-Based Learned Position Compensation}

The second path focuses on motion-mode-dependent drift and refines a RoNIN
baseline trajectory using action-context segments. It learns a two-dimensional
position correction from frozen RoNIN features and predicted carrying-mode
context. For each inference window centered at time
$\tau_i$, the corresponding action segment $a_n$ satisfying
$\tau_i\in[t_n^{\mathrm{st}},t_n^{\mathrm{ed}}]$ provides a five-dimensional
context vector
\begin{equation}
\mathbf{c}_i=
\bar{\mathbf{p}}_n^{\mathrm{act}}
=
\left[
 \bar{p}_n^{\mathrm{hand}},
 \bar{p}_n^{\mathrm{call}},
 \bar{p}_n^{\mathrm{down}},
 \bar{p}_n^{\mathrm{swing}},
 \bar{p}_n^{\mathrm{pocket}}
\right]^\top,
\label{eq:iv_action_context}
\end{equation}
which corresponds to holding, calling, arm-down, arm-swinging, and
pocket modes, respectively. The method carries forward the probability
vector from the segment covering the current timestamp, which helps handle
ambiguous transitions between motion modes smoothly.

The RoNIN backbone first produces a baseline velocity estimate
$\hat{\mathbf{v}}_i^{\mathrm{ron}}$ for each IMU window
$\mathbf{X}^{\mathrm{imu}}_i$:
\begin{equation}
\hat{\mathbf{v}}_i^{\mathrm{ron}}
=
\mathcal{R}(\mathbf{X}^{\mathrm{imu}}_i),
\label{eq:iv_ronin_velocity}
\end{equation}
where $\mathcal{R}(\cdot)$ denotes the frozen RoNIN network. Integrating these
window-wise velocity predictions yields the baseline trajectory
$\mathbf{p}^{\mathrm{ron}}(t)$. In parallel, a 512-dimensional intermediate
backbone feature
\begin{equation}
\mathbf{f}_i=\mathcal{B}(\mathbf{X}^{\mathrm{imu}}_i)
\label{eq:iv_ronin_feature}
\end{equation}
is extracted from the RoNIN front-end and supplied to the learned correction
head.

To inject carrying-mode information, the model applies a lightweight
action-conditioned context gate to the context vector. Let
$\mathbf{W}_q$, $\mathbf{W}_k$, $\mathbf{W}_v$, and $\mathbf{W}_o$ denote the
query, key, value, and output projections, respectively, let $\sigma(\cdot)$
denote the sigmoid function, and let $d_a$ be the context-gate dimension. The
scalar gate response is
\begin{equation}
\alpha_i^{\mathrm{ctx}}=
\sigma
\!\left(
\frac{
\left(
\mathbf{W}_q\mathbf{f}_i
\right)^\top
\left(
\mathbf{W}_k\mathbf{c}_i
\right)
}{\sqrt{d_a}}
\right),
\label{eq:iv_context_gate}
\end{equation}
and the gated action context is obtained by
\begin{equation}
\tilde{\mathbf{c}}_i=
\mathbf{W}_o
\!\left(
\alpha_i^{\mathrm{ctx}}\mathbf{W}_v\mathbf{c}_i
\right).
\label{eq:iv_context_gated}
\end{equation}
The fused representation
\begin{equation}
\mathbf{z}_i=
\left[
\mathbf{f}_i
\;\middle\|\;
\tilde{\mathbf{c}}_i
\right]
\label{eq:iv_fused_feature}
\end{equation}
is then fed to a position-correction head composed of a correction branch and
a confidence gate:
\begin{equation}
\Delta\mathbf{p}_i^{\mathrm{raw}}
=
d_{\max}\tanh
\!\left(
\mathcal{D}(\mathbf{z}_i)
\right),
\quad
g_i=
\mathcal{G}(\mathbf{z}_i)\in[0,1],
\label{eq:iv_delta_gate}
\end{equation}
where $\mathcal{D}(\cdot)$ is the multilayer correction regressor,
$\mathcal{G}(\cdot)$ is the gate network, and $d_{\max}$ limits the magnitude
of the raw correction. The final window-center correction is
\begin{equation}
\Delta\mathbf{p}_i=
g_i\,
\Delta\mathbf{p}_i^{\mathrm{raw}}.
\label{eq:iv_delta_final}
\end{equation}
This gated design keeps the model conservative when the action evidence is
limited or the RoNIN feature already appears reliable, and permits larger
compensation when the motion mode implies stronger pose-induced drift.

During training, the RoNIN backbone is frozen and only the position-correction
head is optimized. After timestamp synchronization and the same SE(2) spatial
alignment used for evaluation, the target correction at the window center is
\begin{equation}
\Delta\mathbf{p}^{\star}_i
=
\mathbf{p}^{\mathrm{gt}}(\tau_i)
-
\mathbf{p}^{\mathrm{ron}}(\tau_i).
\label{eq:iv_pos_target}
\end{equation}
The correction loss is
\begin{equation}
\mathcal{L}_{\mathrm{pos}}
=
\frac{1}{B}
\sum_{i=1}^{B}
\left\|
\Delta\mathbf{p}_i-\Delta\mathbf{p}^{\star}_i
\right\|_2^2
+
\beta
\left\|
\Delta\mathbf{p}_i
\right\|_2^2,
\label{eq:iv_pos_loss}
\end{equation}
where $B$ is the number of training windows in the mini-batch, $\beta$ controls
the correction regularizer, and the second term regularizes the correction
magnitude.

Because the correction head operates on strided windows, the predicted
corrections are sparse in time. To obtain a continuous correction stream,
the window-center sequence $\{\Delta\mathbf{p}_i\}$ is first smoothed by a
Gaussian temporal filter and then linearly interpolated onto the full RoNIN time
axis. The final corrected trajectory is written as
\begin{equation}
\mathbf{p}^{\mathrm{corr}}(t)
=
\mathbf{p}^{\mathrm{ron}}(t)
+
\bar{\Delta\mathbf{p}}(t),
\label{eq:iv_pos_corrected}
\end{equation}
where $\bar{\Delta\mathbf{p}}(t)$ denotes the smoothed and interpolated
position correction. In this formulation, holding, calling, and
arm-down segments tend to correspond to relatively stable device poses, while
arm-swinging and pocket segments usually expose larger orientation
and placement mismatch. By conditioning the
correction head on the full action-probability vector, the method can adapt
its correction magnitude and gating behavior continuously across these
carrying modes while preserving a complete output trajectory.

\section{Experiments and Results}

\subsection{Experimental Setup}

The proposed framework was evaluated using two representative types of smartphone
trajectories, denoted as Route A and Route B. The test trajectories were
collected by multiple volunteers in two
large-scale indoor environments covering areas over $400~\mathrm{m}^2$, and
were paired with LiDAR-SLAM reference trajectories, which were treated as
ground truth for quantitative evaluation. The smartphone logs were recorded
with a Xiaomi 14 Pro. Route A represents an approximately 230~s cross-floor
trajectory type with multi-scene transitions, and mainly examines whether
scene-event cues can constrain and repair the accumulated drift of a step-wise
PDR pipeline, especially around gate, stair, and elevator transitions. Route B
represents an approximately 80~s second-floor trajectory type with multiple
phone-use switches, and focuses on whether action-context cues can improve a
learned inertial trajectory under changing phone carrying modes and motion
patterns.
The sensor modalities and nominal sampling rates used in the test logs are
summarized in Table~\ref{tab:experiment_sensor_config}.

\begin{table}[!htbp]
\caption{Smartphone Sensor Configuration Used in the Experiments}
\label{tab:experiment_sensor_config}
\centering
\footnotesize
\setlength{\tabcolsep}{2pt}
\renewcommand{\arraystretch}{1.08}
\begin{tabular}{@{}p{0.30\columnwidth}p{0.43\columnwidth}p{0.20\columnwidth}@{}}
\toprule
Sensor & Recorded attributes & Sampling rate \\
\midrule
Accelerometer & 3-axis linear acceleration & 100--200 Hz \\
Gyroscope & 3-axis angular velocity & 100--200 Hz \\
Magnetometer & 3-axis magnetic field & 100 Hz \\
AHRS & Pitch, roll, yaw, quaternion & 100--200 Hz \\
Barometer & Pressure & 20 Hz \\
Light & Illuminance & 5--10 Hz \\
GNSS & Position and motion status & 1--2 Hz \\
Sound & RMS and sound pressure level & 1.8--1.9 Hz \\
Wi-Fi & Access-point RSS observations & 0.22--0.50 Hz \\
BLE 4.0 & Beacon RSSI observations & 4.95--6.94 Hz \\
\bottomrule
\end{tabular}
\end{table}

The principal parameters used for semantic filtering and trajectory correction
are summarized in Table~\ref{tab:localization_parameters}.

\begin{table}[!htbp]
\caption{Principal Parameters Used in the Localization Experiments}
\label{tab:localization_parameters}
\centering
\footnotesize
\setlength{\tabcolsep}{4pt}
\renewcommand{\arraystretch}{1.08}
\begin{tabular*}{\columnwidth}{@{\extracolsep{\fill}}p{0.70\columnwidth}p{0.24\columnwidth}@{}}
\toprule
Parameter & Value \\
\midrule
Semantic and elevator confidence threshold, $\tau_q$ & 0.9 \\
Minimum transition-event persistence, $N_{\min}$ & 2 consecutive windows \\
Corridor pull strength, $\lambda$ & 0.6 \\
Corridor heading tolerance, $\varepsilon$ & $20^\circ$ \\
Maximum position correction, $d_{\max}$ & 20.0 m \\
Correction regularization coefficient, $\beta$ & 0.005 (training) \\
Gaussian smoothing parameter, $\sigma$ & 30 \\
\bottomrule
\end{tabular*}
\end{table}

The semantic cues used in all localization experiments are predicted labels
and probabilities generated by the recognition module. The recognition front
end was trained on our self-collected large-scale multi-sensor hierarchical
dataset, which contains 186 recording sequences, 23,193 labeled segments, and
32.09~h of valid recordings. As summarized in
Table~\ref{tab:recognition_frontend_summary}, the model achieves high per-level
and joint accuracy on the held-out test split, indicating that it can
provide reliable semantic inputs for the downstream localization modules.

\begin{table}[!htbp]
\caption{Recognition Performance of the Semantic Front End (\%)}
\label{tab:recognition_frontend_summary}
\centering
\footnotesize
\setlength{\tabcolsep}{4pt}
\renewcommand{\arraystretch}{1.08}
\begin{tabular}{@{}lccccc@{}}
\toprule
Model & \shortstack{Environment\\Accuracy} & \shortstack{Scene\\Accuracy} & \shortstack{Activity\\Accuracy} & \shortstack{Joint\\Accuracy} & \shortstack{Mean\\Macro-F1} \\
\midrule
Ours & 97.54 & 95.24 & 98.06 & 94.35 & 96.83 \\
\bottomrule
\end{tabular}
\end{table}

\begin{table}[!htbp]
\caption{Compared Methods Used in the Experiments}
\label{tab:compared_methods}
\centering
\scriptsize
\setlength{\tabcolsep}{4pt}
\renewcommand{\arraystretch}{0.96}
\begin{tabular}{@{}>{\raggedright\arraybackslash}p{0.18\columnwidth}>{\raggedright\arraybackslash}p{0.78\columnwidth}@{}}
\toprule
Method & Description \\
\midrule
PDR & Step-wise dead reckoning baseline driven by pedestrian motion estimation and heading updates. \\
\midrule
RoNIN & Robust neural inertial navigation baseline used as the main learned trajectory reference. \\
MEIO & Motion-aware inertial odometry baseline with adaptive computation design. \\
RIDI & Double-integration-based inertial baseline representing a classic learning-based IMU pipeline. \\
TLIO & Tight learned inertial odometry baseline for drift-aware trajectory estimation. \\
\midrule
\textbf{Our Work} & Hierarchical semantic interface with scene-constrained PDR correction on Route A and action-aware RoNIN position compensation on Route B. \\
\bottomrule
\end{tabular}
\end{table}

The compared methods are summarized in Table~\ref{tab:compared_methods}.
PDR serves as the conventional dead-reckoning reference, with its step-wise
implementation following the smartphone PDR formulation in SmartPDR
\cite{smartpdr}, while RoNIN, MEIO, RIDI, and TLIO represent representative
learned inertial localization approaches \cite{ronin,meionet,ridi,tlio}. These
baselines were selected to cover classical step-wise dead reckoning, robust
neural inertial navigation, motion-aware efficient odometry, double-integration
learning, and tightly coupled learned inertial filtering. The compared learned
baselines were reproduced using their released open-source implementations and
corresponding public benchmark data or pretrained resources, and were evaluated
using the same input trajectories and evaluation protocol. The semantic outputs
of the recognition module described in Section~III are routed according to the
target backbone: scene-constrained correction for PDR in Route A and
action-aware position compensation for the RoNIN-based trajectory in Route B.

Three metrics are used in the main evaluation. The first metric,
$\mathrm{ATE}$, measures the global absolute deviation of the
aligned trajectory and reflects overall localization accuracy. The
second metric, $\mathrm{RTE}@10\mathrm{m}$, evaluates relative
error over fixed 10~m sub-trajectories and is used to characterize local drift
accumulation. The third metric, $\mathrm{APE}_{\mathrm{median}}$, reports the
median point-wise absolute position error over time and thus reflects the
typical error level with reduced sensitivity to extreme outliers.

For fair comparison, all estimated trajectories were first temporally
synchronized with the LiDAR-SLAM reference and then spatially aligned in a
common planar coordinate frame before metric computation. Spatial alignment
uses a two-dimensional rigid transformation, i.e., rotation and translation in
SE(2), with no scale factor. The same alignment and evaluation procedures were
applied to all methods on both routes. Let
$\mathbf{p}_i$ and $\hat{\mathbf{p}}_i$ denote the ground-truth and aligned
estimated positions at the $i$-th timestamp, respectively. The point-wise
position error is defined as
\begin{equation}
e_i = \left\| \hat{\mathbf{p}}_i - \mathbf{p}_i \right\|_2.
\label{eq:exp_point_error}
\end{equation}
Accordingly, ATE and APE$_{\mathrm{median}}$ are computed by
\begin{equation}
\hspace*{-0.6em}
\mathrm{ATE} =
\sqrt{\frac{1}{N}\sum_{i=1}^{N} e_i^2},
\mathrm{APE}_{\mathrm{median}} =
\mathrm{median}\left(\{e_i\}_{i=1}^{N}\right),
\label{eq:exp_ate_ape}
\end{equation}
where $N$ is the number of aligned samples. For local drift evaluation, let
$(a_j,b_j)$ denote the index pair of the $j$-th reference segment whose length
is 10~m. The corresponding metric is defined as
\begin{equation}
\hspace*{-0.8em}
\mathrm{RTE}@10\mathrm{m} =
\sqrt{
\frac{1}{N_{10}}
\sum_{j=1}^{N_{10}}
\left\|
\left(
\hat{\mathbf{p}}_{b_j} - \hat{\mathbf{p}}_{a_j}
\right)
-
\left(
\mathbf{p}_{b_j} - \mathbf{p}_{a_j}
\right)
\right\|_2^2
},
\label{eq:exp_rte10}
\end{equation}
where $N_{10}$ is the number of 10~m segments.
\subsection{Comparative Experiments and Error Analysis}

Table~\ref{tab:overall_comparison} reports the main quantitative comparison.
Route A evaluates scene-constrained PDR correction on a route with multiple
scene transitions, and Route B evaluates action-aware RoNIN compensation on a
route with multiple phone-use switches.

\begin{table}[!htbp]
\centering
\footnotesize
\caption{Overall Quantitative Comparison on Route~A and Route~B}
\label{tab:overall_comparison}
\scriptsize
\renewcommand{\arraystretch}{0.9}
\begin{tabular*}{\columnwidth}{@{\extracolsep{\fill}}llccc@{}}
\toprule
 & Method & ATE (m) & RTE@10\,m (m) & APE$_{\mathrm{med}}$ (m) \\
\midrule
\multirow{6}{*}{Route A}
& PDR & 17.705 & 12.298 & 7.974 \\
& TLIO & 8.666 & 6.164 & 5.214 \\
& MEIO & 5.537 & 5.293 & 7.593 \\
& RIDI & 7.280 & 6.933 & 4.307 \\
& RoNIN & \underline{3.727} & \underline{4.738} & \underline{1.945} \\
& \textbf{Our Work} & \textbf{2.699} & \textbf{3.765} & \textbf{1.524} \\
\midrule
\multirow{6}{*}{Route B}
& PDR & 18.760 & 10.684 & 22.440 \\
& TLIO & 12.729 & 10.690 & 7.872 \\
& MEIO & 10.391 & 9.000 & 9.905 \\
& RIDI & \underline{4.557} & 5.355 & 10.797 \\
& RoNIN & 6.902 & \underline{2.211} & \underline{6.457} \\
& \textbf{Our Work} & \textbf{3.411} & \textbf{1.707} & \textbf{3.851} \\
\bottomrule
\end{tabular*}
\end{table}

\begin{figure*}[!t]
\centering
\includegraphics[width=\textwidth]{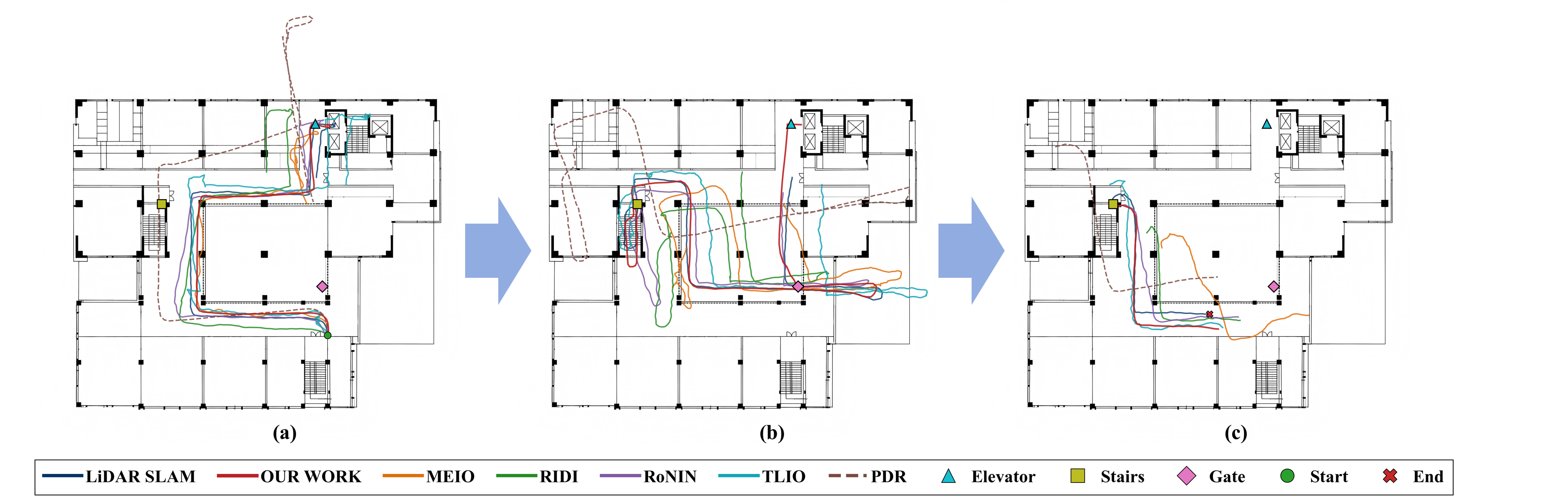}
\vspace{-0.3em}
\caption{Route A trajectory comparison across floor transitions. (a) Initial
trajectory segment on the second floor. (b) Trajectory segment after descending
to the first floor. (c) Trajectory segment after returning to the second floor.}
\label{fig:overall_trajectory_comparison}
\end{figure*}

\begin{figure}[!t]
\centering
\includegraphics[width=\columnwidth]{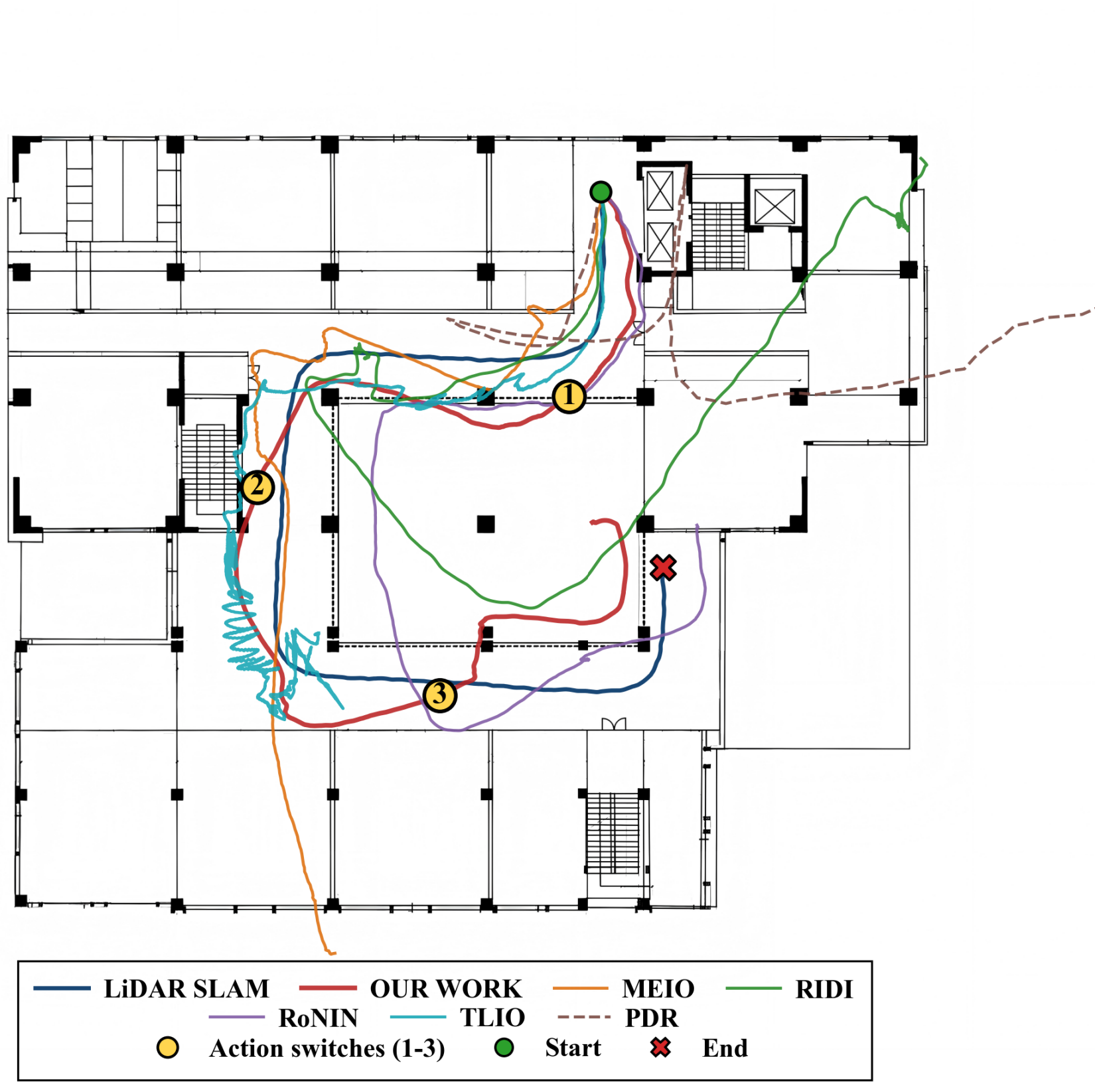}
\caption{Route B overall trajectory comparison under multiple phone-use
switches.}
\label{fig:routeB_overall_trajectory}
\end{figure}

\begin{figure*}[!t]
\centering
\begin{minipage}{0.49\textwidth}
\centering
\includegraphics[width=\linewidth]{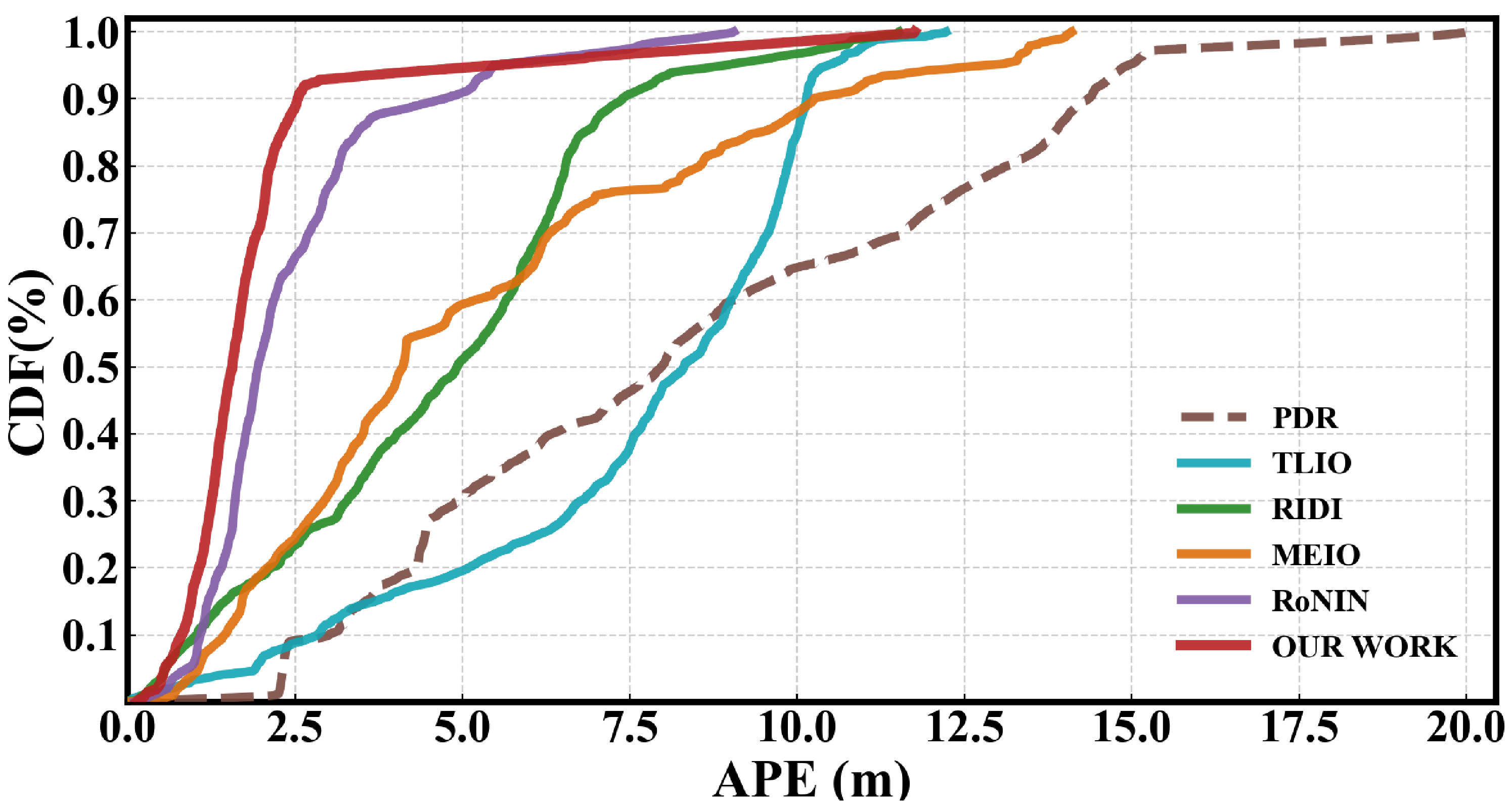}
\vspace{-0.3em}
{\footnotesize (a)\par}
\end{minipage}\hfill
\begin{minipage}{0.49\textwidth}
\centering
\includegraphics[width=\linewidth]{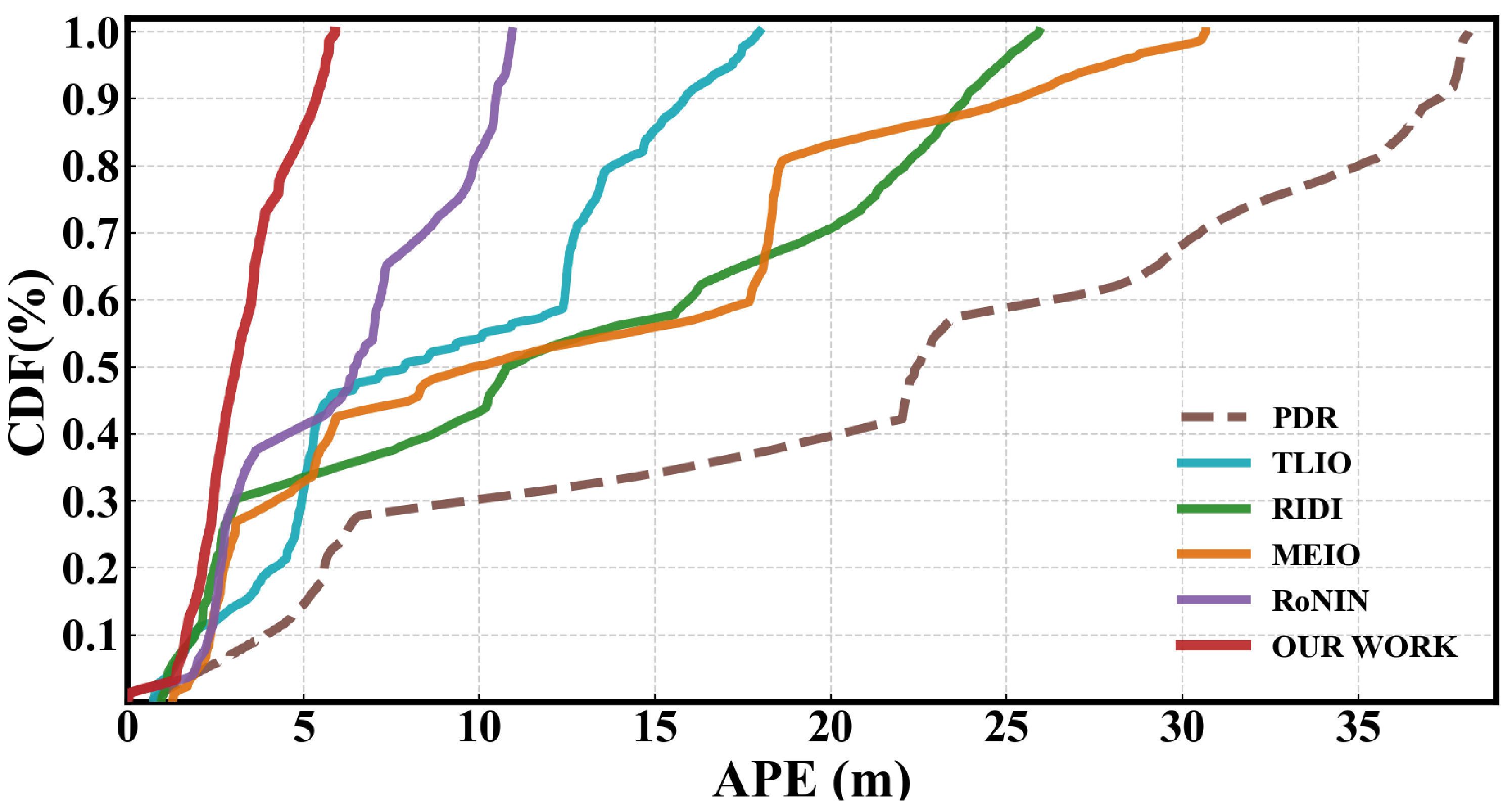}
\vspace{-0.3em}
{\footnotesize (b)\par}
\end{minipage}\\[0.3cm]
\begin{minipage}{0.49\textwidth}
\centering
\includegraphics[width=\linewidth]{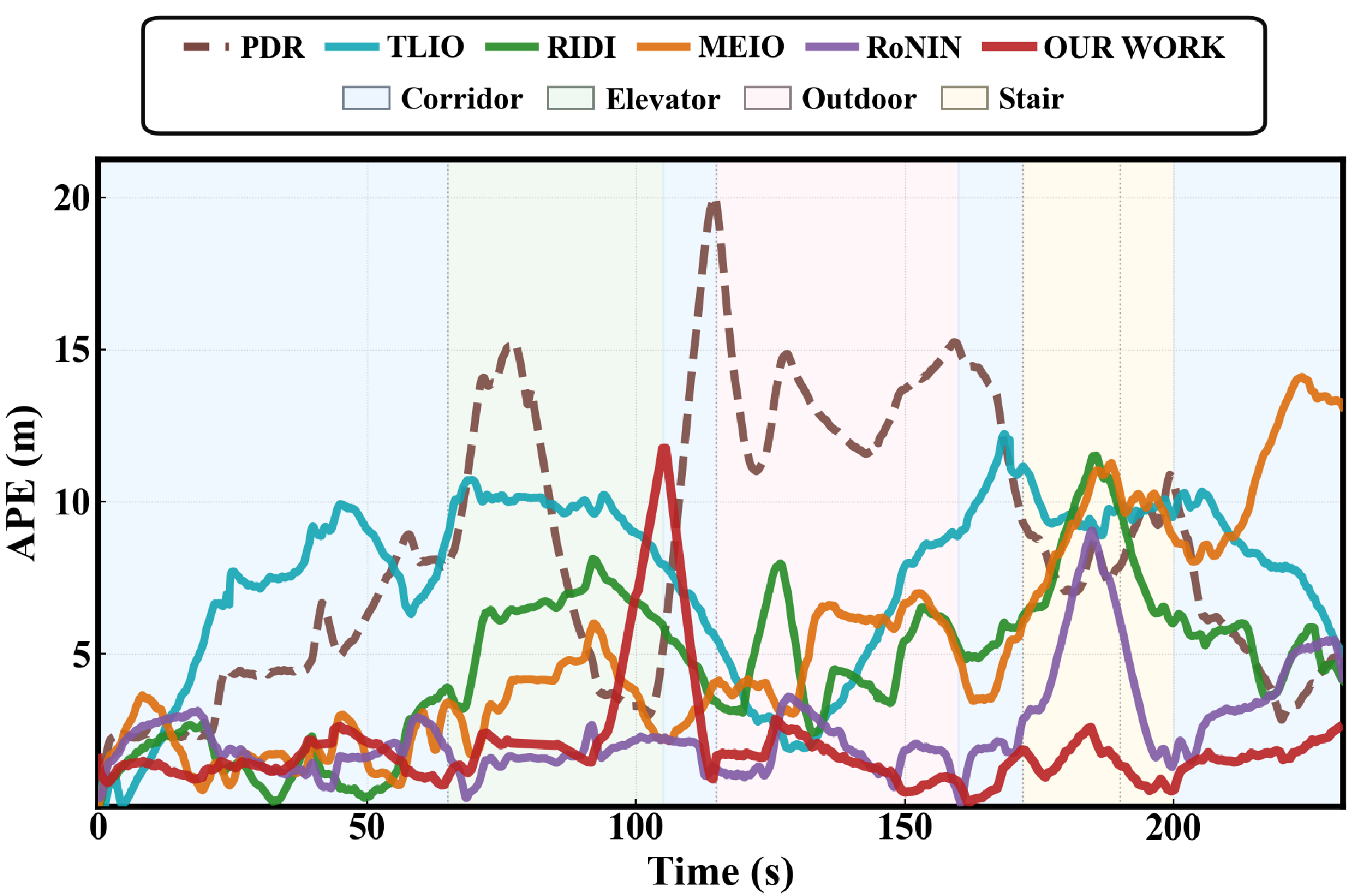}
\vspace{-0.3em}
{\footnotesize (c)\par}
\end{minipage}\hfill
\begin{minipage}{0.49\textwidth}
\centering
\includegraphics[width=\linewidth]{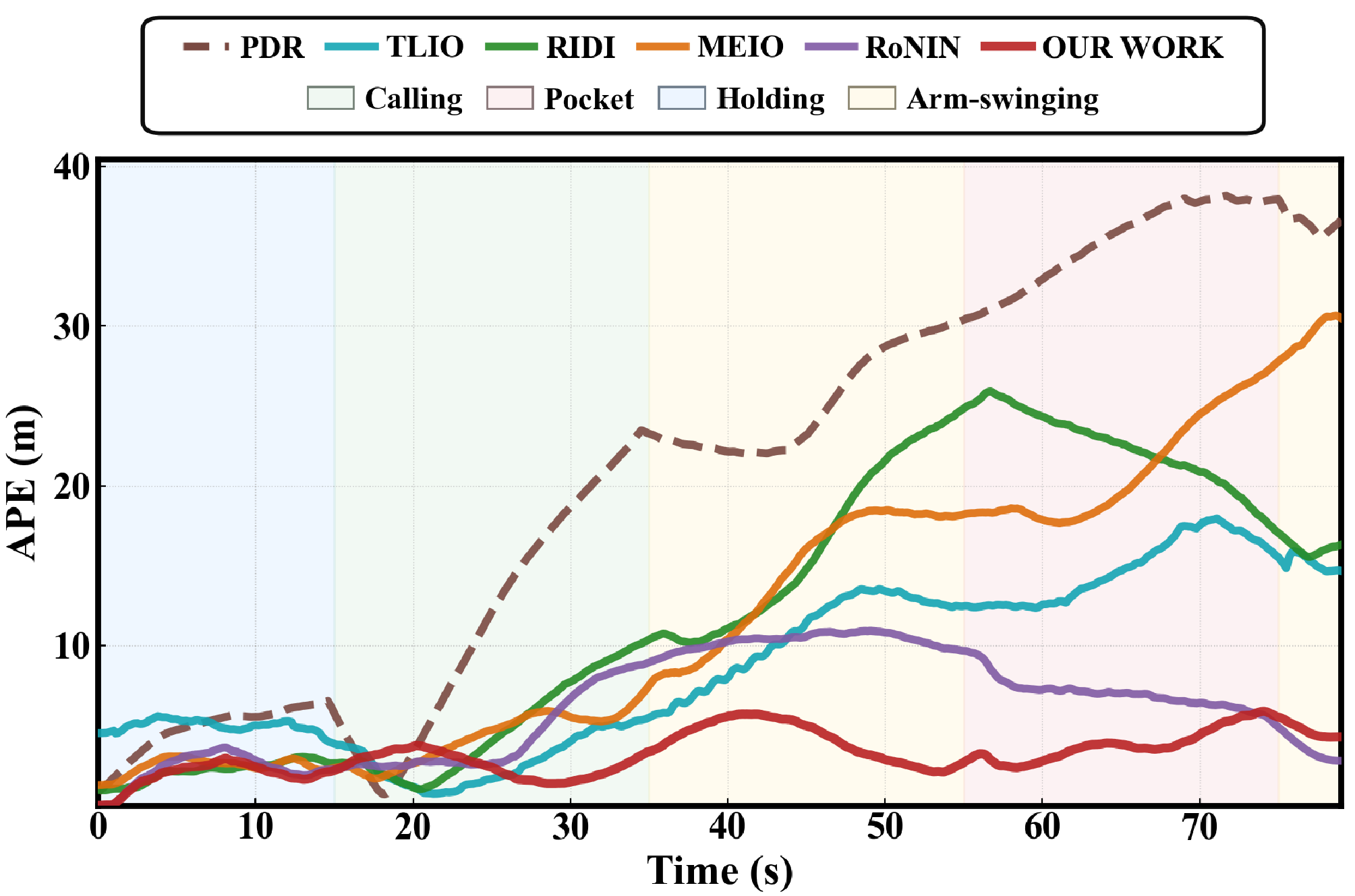}
\vspace{-0.3em}
{\footnotesize (d)\par}
\end{minipage}
\caption{Overall error analysis. The upper row shows displacement-error CDFs,
and the lower row shows APE time series. (a) Route A displacement-error CDF.
(b) Route B displacement-error CDF. (c) Route A APE time-series comparison.
(d) Route B APE time-series comparison.}
\label{fig:overall_error_analysis}
\end{figure*}

\begin{figure*}[!t]
\centering
\begin{minipage}{0.49\textwidth}
\centering
\includegraphics[width=\linewidth]{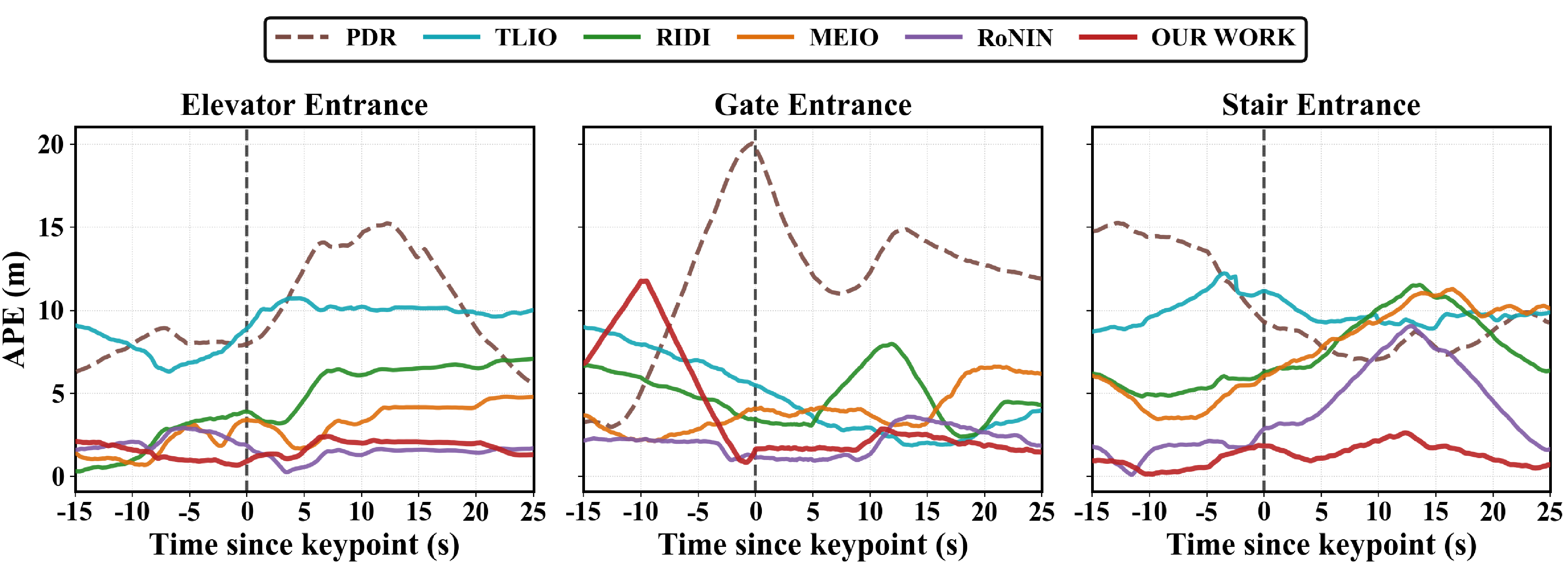}\\[-0.3em]
\footnotesize (a)
\end{minipage}\hfill
\begin{minipage}{0.49\textwidth}
\centering
\includegraphics[width=\linewidth]{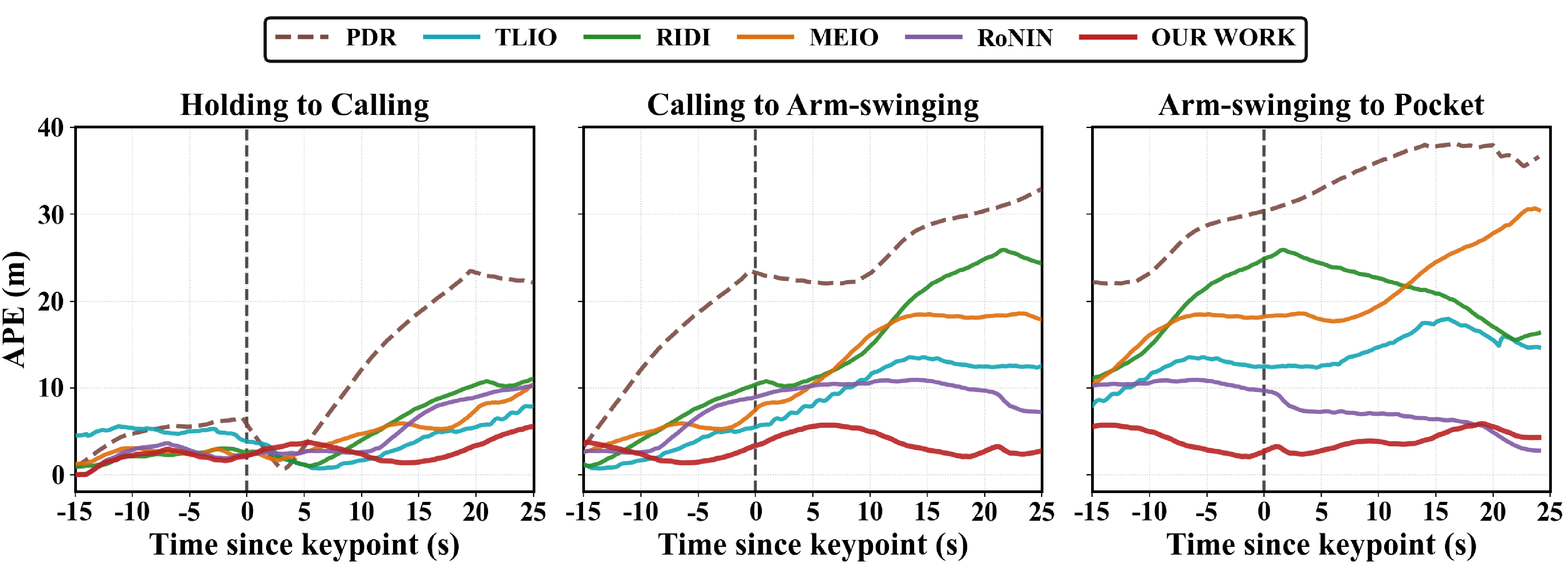}\\[-0.3em]
\footnotesize (b)
\end{minipage}\\[0.3cm]
\begin{minipage}{0.49\textwidth}
\centering
\includegraphics[width=\linewidth]{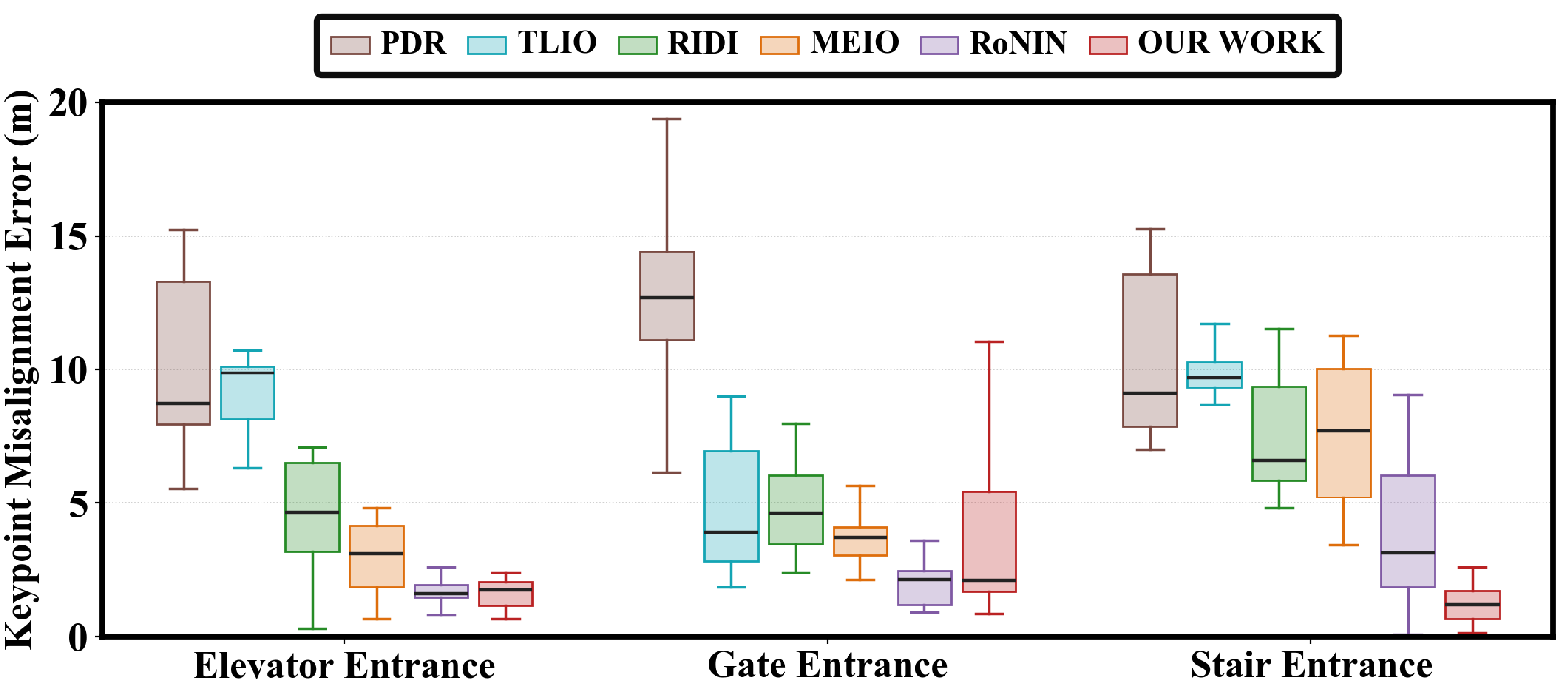}\\[-0.3em]
\footnotesize (c)
\end{minipage}\hfill
\begin{minipage}{0.49\textwidth}
\centering
\includegraphics[width=\linewidth]{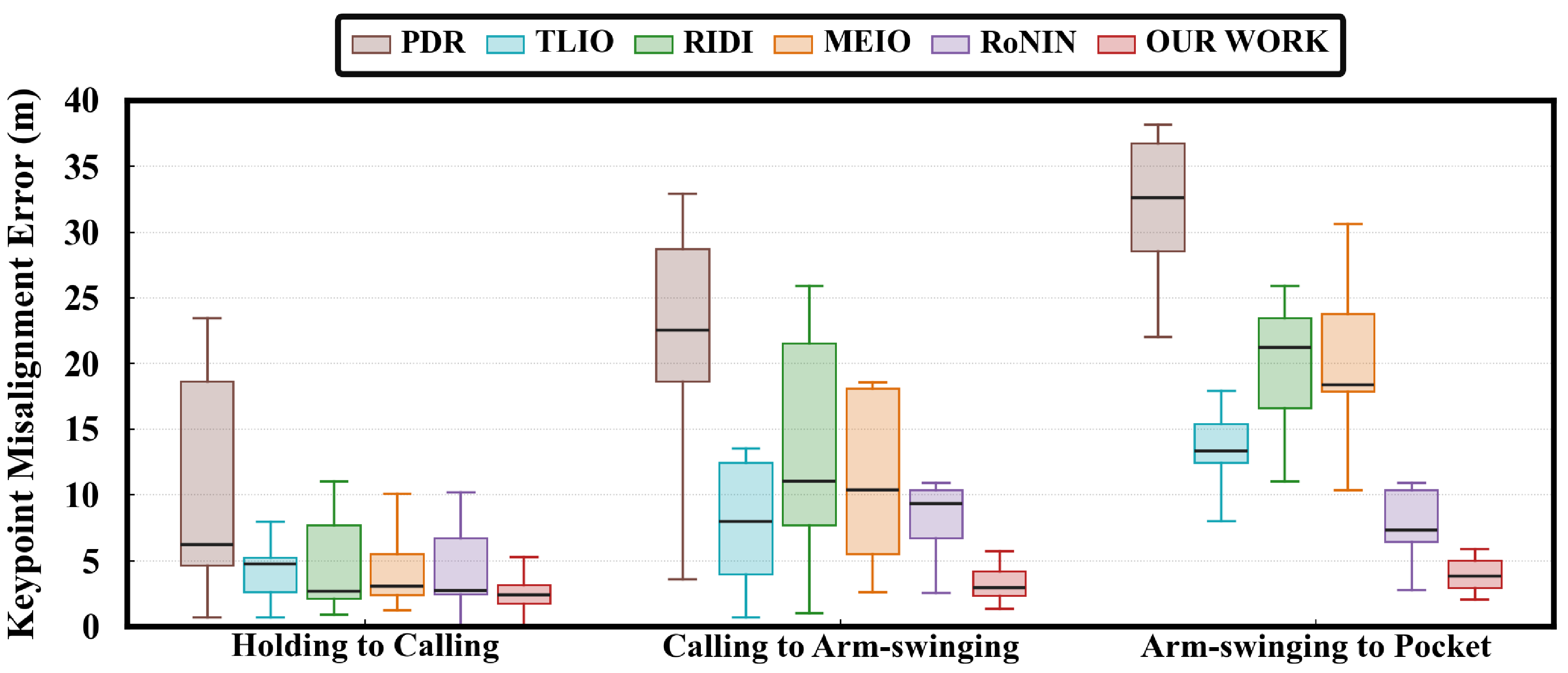}\\[-0.3em]
\footnotesize (d)
\end{minipage}
\caption{Localized gain analysis around semantic triggers. The upper row shows
APE curves in $[-15,25]$~s windows, and the lower row shows APE distributions.
(a) Route A key-point APE evolution. (b) Route B action-switch APE evolution.
(c) Route A key-point APE distribution. (d) Route B action-switch APE
distribution.}
\label{fig:keypoint_gain_analysis}
\end{figure*}

On Route A, the scene-constrained PDR branch reduces ATE from 17.705 to
2.699~m, RTE@10m from 12.298 to 3.765~m, and APE$_{\mathrm{median}}$ from 7.974
to 1.524~m relative to conventional PDR. It also outperforms the learned
baselines on this map-guided, multi-scene route. The three panels in
Fig.~\ref{fig:overall_trajectory_comparison} show that the corrected path stays
closer to the reference across corridor traversal and floor transitions.

Compared with RoNIN on Route B, the proposed method reduces ATE
from 6.902 to 3.411~m, RTE@10m from 2.211 to 1.707~m, and
APE$_{\mathrm{median}}$ from 6.457 to 3.851~m. The corrected trajectory remains
closer to the reference during carrying-mode changes. This result indicates that
the action-context prior stabilizes learned trajectory refinement, as illustrated
in Fig.~\ref{fig:routeB_overall_trajectory}.

Fig.~\ref{fig:overall_error_analysis} further supports the aggregate results. In
the CDF curves of Fig.~\ref{fig:overall_error_analysis}(a) and
(b), our work reaches the same cumulative probability
at smaller APE values than the compared methods, indicating that both the
typical error and the long-tail error are reduced. The time-series curves in
Fig.~\ref{fig:overall_error_analysis}(c) and (d) show that the
proposed method maintains a lower error level over most of the two routes.

On Route A, the corrected trajectory remains stable across the multi-scene route,
except for a short error fluctuation near a transition segment. On Route B, the
advantage becomes clearer after carrying-mode changes, where several baselines
show sustained error growth while our work remains comparatively stable.

\subsection{Gain Around Key Points}

This analysis examines localized improvements around the semantic triggers. For each
structural key point in Route A and each action-switch boundary in Route B, a
fixed time window from $-15$~s to $25$~s is extracted.
The upper row of Fig.~\ref{fig:keypoint_gain_analysis} shows the APE evolution
within these windows, and the lower row summarizes all APE samples collected
from the same windows at a 0.02~s step size. The key points are determined
from the postprocessed recognition outputs after confidence and continuity
filtering.

For Route A, the selected key points correspond to elevator entrances, the gate
entrance, and the stair entrance. The proposed method achieves lower and more
stable APE around the elevator and stair anchors. The gate segment still has a
larger local error, indicating that the improvement is less pronounced near the
gate entrance.

For Route B, the analysis windows are centered on the effective action-switch
boundaries derived from the predicted action sequence, including the transitions
from holding to calling, from calling to arm-swinging, and from arm-swinging to
pocket. The holding-to-calling transition yields only a moderate gain, as the
RoNIN baseline is already stable there. The later carrying-mode boundaries
show clearer separation, especially around the calling-to-arm-swinging and
arm-swinging-to-pocket intervals. This matches the intended role of action-aware
compensation: it is most useful when a phone-use transition causes increased
drift in the learned inertial trajectory.

\subsection{Ablation Study}

\begin{figure*}[!t]
\centering
\scriptsize
\captionof{table}{Ablation results for Route A scene-constrained components and Route B action-aware components.}
\label{tab:ablation_combined}
\setlength{\tabcolsep}{2pt}
\renewcommand{\arraystretch}{1.05}
\begin{tabular*}{0.85\textwidth}{@{\extracolsep{\fill}}
>{\centering\arraybackslash}p{0.10\textwidth}
>{\centering\arraybackslash}p{0.28\textwidth}
>{\centering\arraybackslash}p{0.10\textwidth}
>{\centering\arraybackslash}p{0.12\textwidth}
>{\centering\arraybackslash}p{0.15\textwidth}@{}}
\toprule
Route & Variant & \mbox{ATE (m)} & \mbox{RTE@10m (m)} & \mbox{Median APE (m)} \\
\midrule
\multirow{5}{*}{Route A}
& \textbf{Our Work} & \textbf{2.699} & \textbf{\underline{3.765}} & \textbf{1.524} \\
& w/o Gate Entrance Attraction & 3.753$\uparrow$ & 4.583$\uparrow$ & 3.182$\uparrow$ \\
& w/o Corridor Heading Constraint & \underline{2.747}$\uparrow$ & 3.820$\uparrow$ & \underline{1.617}$\uparrow$ \\
& w/o Backtracking Smoothing & 3.399$\uparrow$ & 4.452$\uparrow$ & 2.219$\uparrow$ \\
& w/o Vertical-Transition Handling & 2.828$\uparrow$ & 2.063$\downarrow$ & 2.142$\uparrow$ \\
\midrule
\multirow{5}{*}{Route B}
& \textbf{Our Work} & \textbf{3.411} & \textbf{1.707} & \textbf{3.851} \\
& w/o Holding & 4.757$\uparrow$ & 4.536$\uparrow$ & 4.103$\uparrow$ \\
& w/o Calling & \underline{4.001}$\uparrow$ & \underline{3.533}$\uparrow$ & 4.062$\uparrow$ \\
& w/o Arm-Swinging & 4.339$\uparrow$ & 4.221$\uparrow$ & 3.431$\downarrow$ \\
& w/o Pocket & 4.276$\uparrow$ & 4.170$\uparrow$ & \underline{3.707}$\downarrow$ \\
\bottomrule
\end{tabular*}

\vspace{6pt}
\begin{minipage}{0.90\textwidth}
\centering
\begin{minipage}[b]{0.47\linewidth}
\centering
\includegraphics[width=\linewidth]{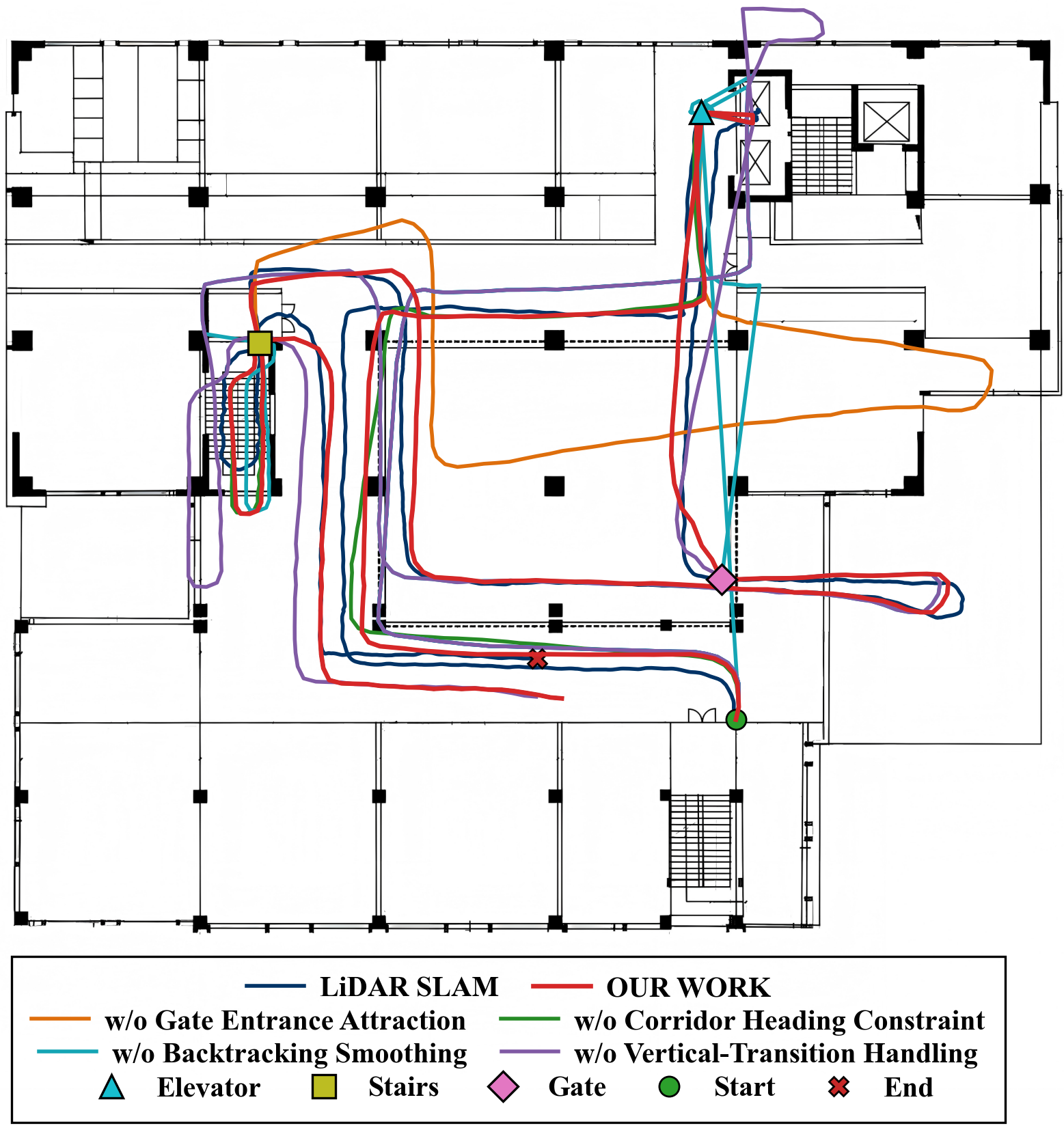}\\[-0.3em]
\footnotesize (a)
\end{minipage}\hfill
\begin{minipage}[b]{0.48\linewidth}
\centering
\includegraphics[width=\linewidth]{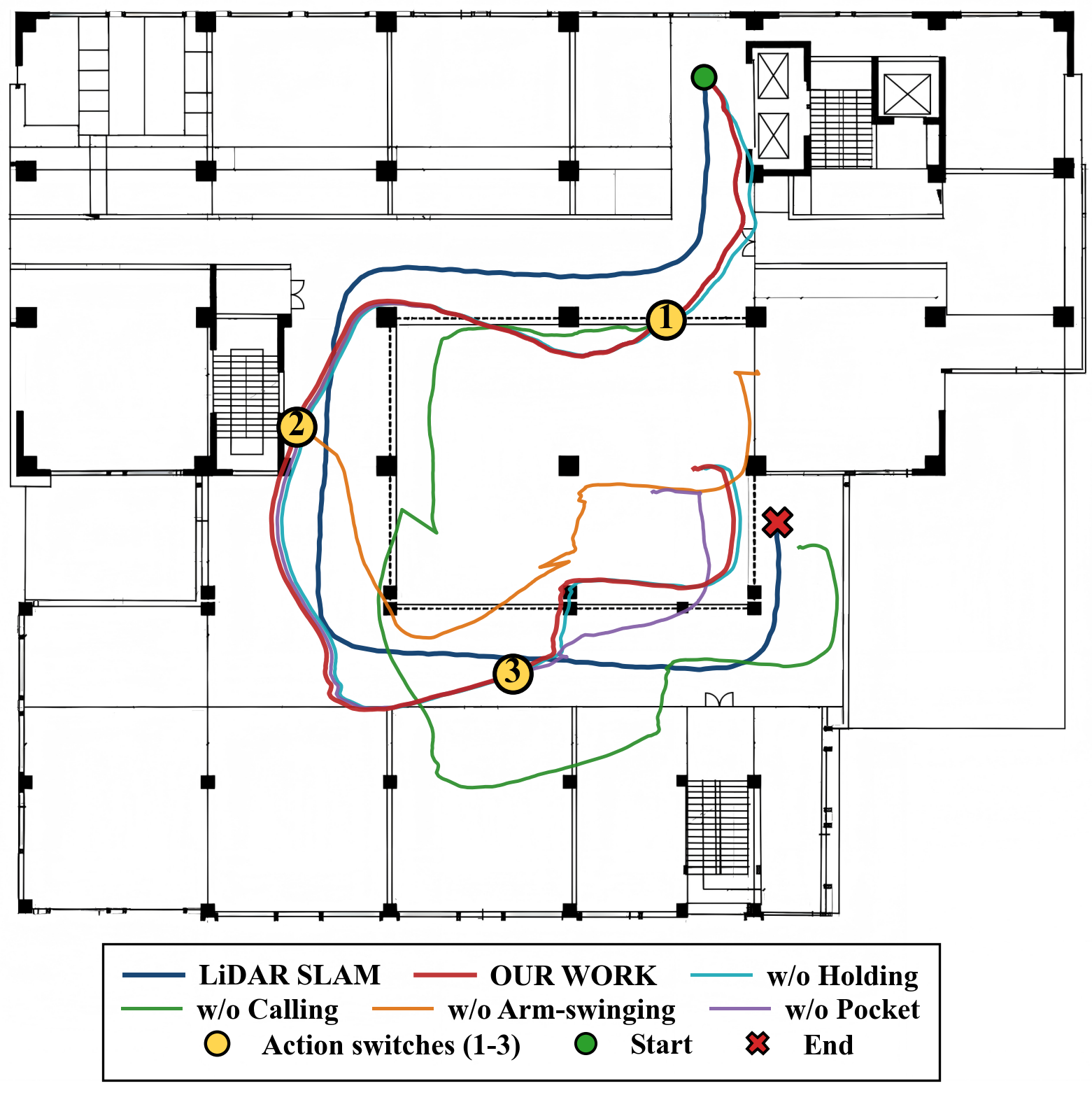}\\[-0.3em]
\footnotesize (b)
\end{minipage}
\end{minipage}
\captionof{figure}{Ablation trajectory comparison. (a) Route A removes one
scene-constrained PDR component at a time. (b) Route B removes one action-aware
compensation cue at a time.}
\label{fig:ablation_trajectory}
\end{figure*}

To isolate component effects, one component or semantic cue is removed at a
time while all other settings remain unchanged. For Route A, each ablation
disables one scene-constrained PDR component: gate entrance attraction, corridor
heading constraint, backtracking smoothing, or vertical-transition handling.
For Route B, each ablation suppresses the action-aware compensation cue for one
action mode appearing in the route.
Table~\ref{tab:ablation_combined} reports the quantitative changes, and
Fig.~\ref{fig:ablation_trajectory} visualizes the corresponding trajectory
deviations.

For Route A, removing the gate entrance attraction term causes the largest overall
degradation. The ATE rises from 2.699 to 3.753, the RTE@10m rises from 3.765
to 4.583, and the median APE more than doubles from 1.524 to 3.182. Removing
backtracking smoothing also degrades all three metrics, which confirms that event-driven
correction alone is insufficient without propagating the recovered alignment to
neighboring segments. Removing the corridor heading constraint produces only a
mild increase, indicating that corridor heading guidance mainly acts as a stabilizing
prior. The vertical-transition handling component shows a different pattern:
removing it increases ATE and
APE$_{\mathrm{median}}$ to 2.828 and 2.142, even though the local RTE@10m
becomes smaller. Combined with the trajectory overlay in
Fig.~\ref{fig:ablation_trajectory}, this suggests that vertical-transition handling is
most valuable for preserving global consistency around elevator-related
transitions.

For Route B, all four action-mode cues contribute to route-level
consistency. Removing the holding cue yields the largest degradation, with
ATE increasing from 3.411 to 4.757 and RTE@10m from 1.707 to 4.536. The
calling, arm-swinging, and pocket cues also cause clear degradation in
ATE and RTE@10m when removed, and the trajectory overlay reveals more
pronounced deviation after the action-switch regions. The median APE does not
increase monotonically for every reduced variant; for example, the
\mbox{w/o Arm-Swinging} variant has a slightly smaller APE$_{\mathrm{median}}$
than the complete system. This behavior indicates that the action-mode cues
mainly suppress route-level drift and switch-boundary distortion, while the
point-wise median error alone does not fully capture their contribution.

\section{Conclusion}
This study contributes a hierarchical semantic enhancement framework for
smartphone indoor localization that connects multi-sensor context recognition
with a scene-constrained PDR branch and an action-aware RoNIN compensation
branch. The results highlight the value of using hierarchical context as a semantic interface
between recognition outputs and trajectory correction. In particular, scene cues
can express structure-related opportunities for constraining PDR trajectories,
while activity cues can identify phone-use changes that are relevant to
learned drift compensation. These observations also show that the effect of
semantic enhancement depends on whether the semantic cues are reliable,
temporally stable, and aligned with the localization errors to be corrected.

Several limitations should be acknowledged. The scene-constrained PDR branch
relies on prior map information, including corridor directions and landmark
locations; incomplete or outdated maps may therefore weaken its corrections,
particularly in dynamically changing indoor spaces. In addition, the evaluation
covered only two route types in two indoor environments using a single smartphone
model. Generalization across buildings, devices, users, and longer trajectories
requires further validation.

Future work will investigate online map adaptation and landmark discovery to
reduce dependence on map priors, while extending the evaluation to more diverse
conditions. More adaptive semantic triggering will also be explored to account
for uncertainty and temporal alignment. With broader validation, the framework
may support pedestrian navigation and tracking in multi-floor public buildings,
transportation hubs, indoor construction sites, and emergency-response
environments, where structural transitions and phone-use variations commonly
affect localization reliability.

\end{document}